\documentclass[letterpaper, 10 pt, journal, twoside]{IEEEtran}
\usepackage{amsmath,amsfonts,amssymb}
\usepackage{algorithmic}
\usepackage{algorithm}
\usepackage{array}
\usepackage[caption=false,font=normalsize,labelfont=sf,textfont=sf]{subfig}
\usepackage{textcomp}
\usepackage{stfloats}
\usepackage{url}
\usepackage{verbatim}
\usepackage{graphicx}
\usepackage{cite}
\def\BibTeX{{\rm B\kern-.05em{\sc i\kern-.025em b}\kern-.08em
    T\kern-.1667em\lower.7ex\hbox{E}\kern-.125emX}}
\usepackage{booktabs}
\usepackage{multirow}
\usepackage{siunitx}
\usepackage{xcolor}
\usepackage{color,soul}
\sethlcolor{yellow}
    
\usepackage[pdfstartview=XYZ,
bookmarks=true,
colorlinks=true,
linkcolor=black,
urlcolor=black,
citecolor=black,
bookmarks=true,
linktocpage=true, 
hyperindex=true
]{hyperref}

\usepackage{orcidlink}

\begin{document}

© 2022 IEEE. Personal use of this material is permitted.
Permission from IEEE must be obtained for all other uses,
including reprinting/republishing this material for advertising
or promotional purposes, collecting new collected works
for resale or redistribution to servers or lists, or reuse of
any copyrighted component of this work in other works.
This work has been submitted to the IEEE for possible
publication. Copyright may be transferred without notice,
after which this version may no longer be accessible.

Please cite the newer, accepted version that has the DOI below:
Digital Object Identiﬁer 10.1109/TCST.2023.3338112

\title{Physical Human-Robot Interaction Control of an Upper Limb Exoskeleton with a Decentralized Neuro-Adaptive Control Scheme}

\author{Mahdi Hejrati \orcidlink{0000-0002-8017-4355}, Jouni Mattila \orcidlink{0000-0003-1799-4323}
\thanks{ The TITAN (Teaching human-like abilities to heavy mobile manipulators through multisensory presence) project is funded by the Technology Industries of Finland Centennial Foundation and the Jane and Aatos Erkko Foundation Future Makers programme. 2020-2023. (corresponding author: Mahdi Hejrati) }
\thanks{ Mahdi Hejrati is with the Department of Engineering and Natural Science, Tampere University, 7320 Tampere, Finland (e-mail: mahdi.hejrati@tuni.fi).}
\thanks{ Jouni Mattila is with the Department of Engineering and Natural Science, Tampere University, 7320 Tampere, Finland (e-mail: Jouni.Mattila@tuni.fi).}
\thanks{Digital Object Identifier (DOI): 10.1109/TCST.2023.3338112.}
}

\markboth{IEEE Transactions}%
{Hejrati and Mattila: PHRI Control of an Upper Limb Exoskeleton with a Decentralized Neuro-Adaptive Control Scheme} 


\maketitle

\begin{abstract}
Within the concept of physical human-robot interaction (pHRI), the paramount criterion is the safety of the human operator interacting with a high-degree-of-freedom (DoF) robot. Consequently, there is a substantial demand for a robust control scheme to establish safe pHRI and stabilize nonlinear, high DoF systems. In this paper, an adaptive decentralized control strategy is designed to accomplish the abovementioned objectives. A human upper limb model and an exoskeleton model are decentralized and augmented at the subsystem level to enable decentralized control action design. Human exogenous torque (HET), which can resist exoskeleton motion, is estimated using radial basis function neural networks (RBFNNs). Estimating human upper limb and robot rigid body parameters, as well as HET, makes the controller adaptable to different operators, ensuring their physical safety during the interaction. To guarantee both safe operation and stability, the barrier Lyapunov function (BLF) is utilized to adjust the control law. This study also considers unknown actuator uncertainties and constraints to ensure a smooth and secure pHRI. Additionally, it is shown that incorporating RBFNNs and BLF into the original VDC improved its performance. The asymptotic stability of the entire system is established through the concept of virtual stability and virtual power flows (VPFs) under the proposed robust controller. Experimental results are presented and compared with those obtained using PD and PID controllers to showcase the robustness and superior performance of the designed controller, particularly in controlling the last two joints of the robot.
\end{abstract}


\section{Introduction}
\subsection{Overview}
\IEEEPARstart{P}{hysical} human-robot interaction (pHRI) combines human and robotic capabilities through physical interaction to carry out an assigned task. The typical application modes of such an interaction are assistance, cooperation, and teleoperation. Exoskeletons are wearable robots that can work in both assistance mode and teleoperation mode. In assistance mode, pHRI can be either for short-term activities, such as performing daily tasks, or for more prolonged activities, such as a power extender \cite{kazerooni1991dynamics}. In contrast, in teleoperation mode, exoskeletons are utilized as master robots that are worn by the operator, sending and receiving commands to a slave robot. Therefore, pHRI persists until the task on the slave side is completed \cite{rebelo2014bilateral}. These applications require close, safe, and dependable physical interactions between humans and robots within a shared workspace \cite{haddadin2016physical}. The overarching research objective in safe pHRI has three categories: interaction safety assessment, interaction safety through design, and interaction safety through planning and control \cite{pervez2008safe}. In this paper, we focused on the safety aspect of pHRI through control and designed a decentralized control scheme that is aimed at not only ensuring safety but also achieving precise control objectives.

\subsection{Related Works}
\subsubsection{Exoskeleton Control}
In the pHRI concept, exoskeletons can work in active or passive assistance mode. In passive assistance mode, the goal of the exoskeleton is to track the desired trajectory while supporting a human limb. In the active assistance mode, the exoskeleton helps humans to accomplish a task while compensating for part of the required force \mbox{\cite{zhang2020integrated}}. Both approaches have been widely examined in the literature for task space \cite{sharifi2021impedance,li2018physical} and joint space \cite{brahmi2018passive}. In \cite{li2015nonlinear} a nonlinear disturbance observer-based controller with a fuzzy approximation is designed to accomplish control objectives in the presence of model uncertainties and input constraints. In \cite{brahmi2018adaptive}, dynamic uncertainty has been handled utilizing the time delay estimation (TDE) method, and an adaptive backstepping control method has been employed to achieve control goals. Radial basis function neural networks (RBFNNs) along with an adaptive backstepping sliding mode control (SMC) method have been utilized in \cite{wu2019rbfn} to address uncertainty and achieve control objectives. The compliant control of an upper limb exoskeleton based on SMC has been investigated in \cite{brahmi2018compliant}. Moreover, in \cite{brahmi2019adaptive}, integral second-order SMC has been exploited for the control problem of upper limb exoskeletons. Although the abovementioned control schemes have shown good performance, they are difficult to implement. For example, backstepping can result in an explosion of complexity, even for a simple system \mbox{\cite{yip1998adaptive}}. Additionally, the mentioned works are based on the Lagrangian dynamics models of robots. Notably, these methods entail a level of computational complexity (computational burden) that scales proportionally to the fourth power of the degrees of freedom (DOFs) \mbox{\cite{zhu2010virtual}}. It can be expected that new subsystem-dynamics-based control design methods that are grounded in Newton-Euler (NE) dynamics will gain more popularity \mbox{\cite{mattila2017survey}}. One of these emerging NE-based control approaches is known as virtual decomposition control (VDC), as introduced in \mbox{\cite{zhu2010virtual}}. VDC offers the advantage of a computational burden proportional to the number of subsystems rather than the fourth power of DOFs in motion. Furthermore, the VDC approach does not suffer from the explosion of complexity because it breaks down the entire complex system into subsystems and focuses on controller design and stability analysis at the subsystem level. In light of these factors, to ensure the safe pHRI of high-DoF robots, such as the 7 DoF upper limb exoskeleton in this study (Fig. 1), a decentralized control scheme is designed by employing VDC. A comprehensive explanation of VDC is presented in Section II.

\subsubsection{Safe pHRI Control}
To guarantee interaction safety through control during pHRI, several factors must be taken into account in the controller design. The first issue is actuator input constraints, for instance, saturation, deadzone, or backlash. Actuator constraints widely exist in robotic systems and can destroy control performance and result in instability and unsafe pHRI \cite{zhou2006adaptive}. In \cite{sun2012saturated} a saturated adaptive robust controller is designed to handle possible actuator saturation and model uncertainties. However, one well-known approach to address these constraints is exploiting RBFNNs. Due to their universal function approximation ability, RBFNNs are mostly utilized to estimate unknown functions in robotics systems. In \cite{liu2015adaptive}, RBFNNs have been employed to estimate deadzone. In \cite{li2016adaptive}, RBFNNs along with an auxiliary term have been introduced to handle input saturation. 

The second factor that can affect the safety of pHRI is the accuracy of the represented dynamics of the human limb and robot. Due to the existence of unmodeled dynamics in the robot model, especially in the actuator part, operating at high frequency can excite those unmodeled dynamics and result in instability and unsafe pHRI. Numerous works have been performed to address unmodeled dynamics in robotic systems \cite{yu2022neural,liu2002decentralized}. In addition to robot dynamics, human limb dynamics must be considered in the pHRI model. In \cite{haninger2018identification}, the impact of human dynamics on pHRI has been investigated, and its importance has been shown. To consider human upper limb dynamics in interaction with exoskeletons, a number of challenges must be overcome, e.g., estimating human exogenous torques (HET) and deriving inertia, damping, and stiffness matrices for a second-order dynamic model. Different methods have been proposed to estimate HETs. In \cite{malysz2011kinematic} the HET has been estimated using a fast parameter adaptation function. In \cite{lampinen2021force}, HET has been expressed in linear-in-parameter form, and the unknown constant part was estimated by a projection function. While VDC is fundamentally a model-based controller, complete knowledge of the system is not always accessible. A primary objective of this study is to address the aforementioned challenges by integrating RBFNNs into the control law. This integration enhances the safety of physical human-robot interaction (pHRI) and allows us to retain the advantageous characteristics of the NE-based controller, all the while considering the pHRI model in the control design and effectively addressing system constraints and uncertainties. 

The third challenge for safety in pHRI is state constraints. Although the mechanical range of exoskeletons is achievable for human operators, some postures can cause harm to operators due to fast robot movement, applied high torque from robot to operator, or high tracking error. To take this constraint into account, the barrier Lyapunov function (BLF) with tangent, cosecant, and logarithm functions is utilized in the literature \cite{he2015adaptive,rout2020modified,wu2019reference}. In this work, for the first time in the context of the VDC approach, the logarithmic BLF is utilized to modify the control law to handle state constraints. This required the incorporation of BLF into the virtual stability concept while keeping the decentralized feature of the control scheme, which increased the complexity of the stability analysis.

\begin{figure}[t]
      \centering
      \subfloat[]{\includegraphics[width = 0.2\textwidth]{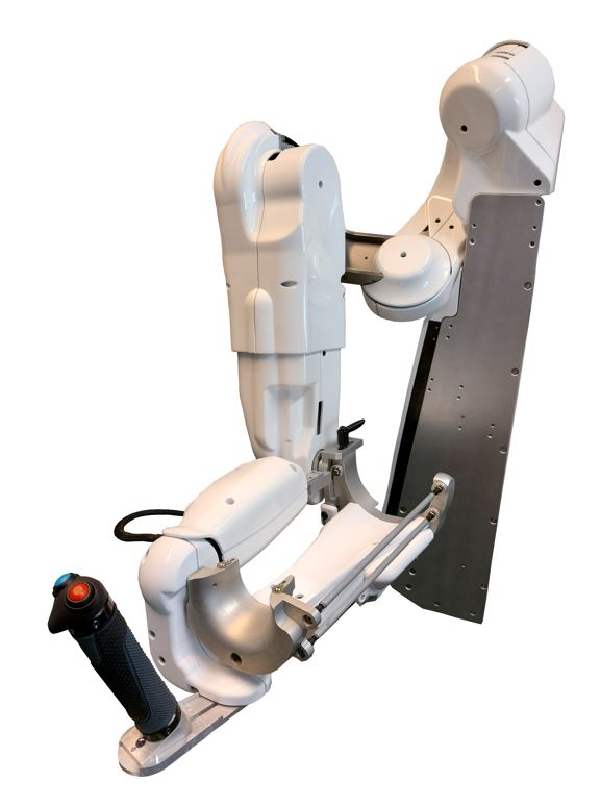}
      \centering
      \label{fIg4}}
      \hfil
      \subfloat[]{\includegraphics[width = 0.22\textwidth]{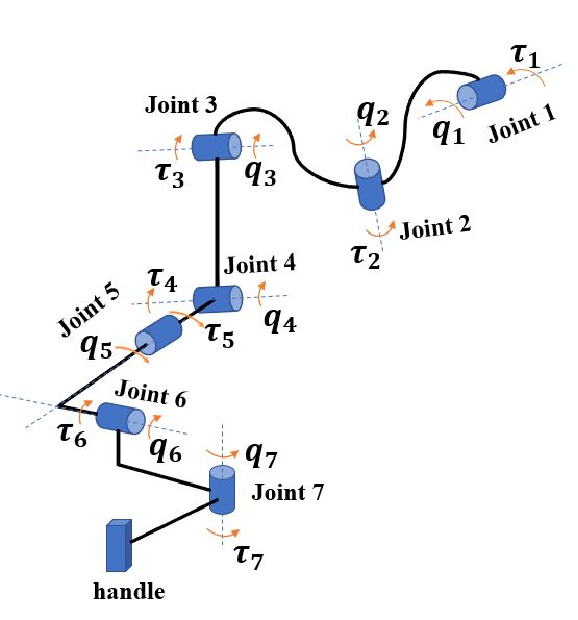}
      \centering
      \label{fIg5}}
      \caption{ a) Upper limb exoskeleton utilized in this study for pHRI, b) joints configuration of the exoskeleton}
   \end{figure}
   
\subsection{Aims and Contributions}

The primary objective of this paper is to employ VDC as a decentralized control scheme to control pHRI amid the presence of unknown, unmodeled dynamics and input constraints. To ensure a safe pHRI, we consider and address state constraints by exploiting the BLF. Additionally, for a more precise pHRI model and enhanced safety, we estimate both HET and human arm characteristics, including mass, damping, and stiffness matrices, at the subsystem level. RBFNNs are also employed to tackle unknown combined input constraints and unmodeled dynamics. To increase the robustness of the controller, the unknown disturbance is considered and estimated by RBFNNs. Notably, each of these contributions is being made for the first time in the context of the VDC approach. These contributions necessitated modifications in the modeling phase and addressing complexities in stability analysis while upholding the decentralized nature of the VDC approach. The contributions of this paper are expressed as follows:

\begin{itemize}
    \item An augmented human-robot model is derived and utilized for pHRI control. The proposed model represents the interaction model in joint space.
    \item The exoskeleton robot in this study, ABLE \cite{garrec2020design}, which is demonstrated in Fig. 1, has a complex forearm and wrist mechanism composed of a belt, ball screw, and cable. In order to handle the unmodeled dynamics and input constraints in such actuators, RBFNNs are employed. Additionally, HET in the human-robot augmented model is estimated by RBFNNs. These increased the robustness of the VDC controller.
    \item BLF is exploited to apply state constraints to the system. This ensures the physical safety of pHRI as well as asymptotic stability by keeping the joints inside the predefined safe region. The incorporation of these contributions into VDC not only resulted in a robust VDC but also improved its performance. The asymptotic stability of the overall system in the presence of the presented controller is proven by means of the virtual stability concept and experimentally validated.
\end{itemize}

The rest of the paper is organized as follows. Section II expresses the fundamental mathematics of the VDC approach along with the essential lemmas and definitions utilized in this paper. Section III describes the procedure of decentralization and augmentation of human upper limb dynamics with robot dynamics at the subsystem level. Control design and stability analysis are performed in Section IV. Experimental results are provided in Section V, and the validity of the stability and control results is verified. Finally, Section VI concludes this study.

\section{Mathematical Preliminaries}
\subsection{Virtual Decomposition Control}

\emph{Virtual decomposition control} (VDC) \cite{zhu2010virtual} keeps the overall robotic system asymptotically stable by using subsystem dynamics (link connections and joints). The use of a scalar term called the virtual power flow (VPF), which is defined as the inner product of the velocity error and the force error in a common frame, is one element of this method. The VPF is introduced at each virtual cutting point (VCP), where a virtual "disconnection" is placed. The VPF serves as a distinctive definition of the dynamic interaction between subsystems and is essential in defining the virtual stability of each subsystem. Thanks to VCP and VPF, the entire intricate system, comprising multiple robots engaged in manipulating common objects while interacting with an environment, can be decomposed into subsystems. Each subsystem consists of a rigid body and actuator dynamic parts. By doing so, the design of control action and stability analysis for the complex system is converted to design a local controller at each subsystem level \mbox{\cite{zhu2010virtual}}. The local control law, to be elaborated upon later, has two terms to consider both dynamics: the first term compensates for the dynamics of the rigid body, such as gravitational and inertia effects, while the second term computes the required action to achieve the control goal of the subsystem. Each term is separately computed, and, therefore, the nonlinearity of the coupled rigid body and actuator dynamics is avoided in designing the local control law while the nonlinearity of the entire system is considered in the overall control law. As a result, this method is specifically designed for controlling complex systems, with many significant state-of-the-art control performance improvements with robotic systems \cite{mattila2017survey,koivumaki2015stability,zhu2000stability}.

Despite all the above-mentioned advantages of VDC, it suffers from some drawbacks. Since the VDC approach deals with the decomposed model, incorporating existing methods such as time-varying BLF and prescribed performance methods to VDC is not as easy as for other methods, such as backstepping. In the VDC model for the hydraulic manipulator, it is even difficult to incorporate constant BLF. Additionally, it uses a projection function for parameter estimation, which requires 13\mbox{\(n\)} adaptation gain and 26\mbox{\(n\)} upper and lower parameter bounds, with \mbox{\(n\)} being the number of rigid bodies. In cases of high DoF systems, providing such a long list of parameters is arduous. To remove the projection function, in \mbox{\cite{hejrati2022decentralized}}, we proposed a novel VDC scheme that exploits the natural adaptation law (NAL) function to estimate unknown parameters. The NAL function only requires one adaptation gain for the entire system and removes upper and lower bounds while ensuring a physical consistency condition. Although the proposed method in \mbox{\cite{hejrati2022decentralized}} and this study eliminates the mentioned problem for rigid body parts and actuators with electric motors, the problem of hydraulic actuators is still unsolved and requires providing numerous parameters. In \cite{luna2016virtual}, the original VDC approach \cite{zhu2010virtual} is utilized to control a 7 DoF exoskeleton with a projection function for parameter estimation. Additionally, the authors investigated only the joint tracking control problem, whereas in this paper, the goal is to obtain a safe pHRI considering a complex human-robot interaction model. 

\subsection{VDC Foundations}
Consider \{B\} as a frame that is attached to a rigid body. Then, the 6D linear/angular velocity vector \(^B\mathcal{V}\in \Re^6\) and force/moment vector \(^B\mathcal{F}\in \Re^6\) can be expressed as follows \cite{zhu2010virtual}:

\begin{equation*}
^B\mathcal{V} = [^Bv,\,^B\omega]^T,\quad ^B\mathcal{F} = [\,^Bf,\,^Bm]^T
\end{equation*}
where \(^Bv\in \Re^3\) and \(^B\omega\in \Re^3\) are the linear and angular velocities of frame \{B\}, and \(^Bf\in \Re^3\) and \(^Bm\in \Re^3\) are the force and moment expressed in frame \{B\}, respectively. The transformation matrix that transforms force/moment vectors and velocity vectors between frames \{B\} and \{T\} is \cite{zhu2010virtual},

\begin{equation}\label{equ1}
^BU_T = \begin{bmatrix}
^BR_T & \textbf{0}_{3\times3} \\
(^Br_{BT}\times)\, ^BR_T & ^BR_T
\end{bmatrix}
\end{equation}
where \(^BR_T \in \Re^{3\times3} \) is a rotation matrix between frame \{B\} and \{T\}, and (\(^Br_{BT}\times\)) is a skew-symmetric matrix operator defined as,
\begin{equation}\label{equ2}
^Br_{BT}\times = \begin{bmatrix}
0 & -r_z & r_y \\
r_z & 0 & -r_x\\
-r_y & r_x & 0
\end{bmatrix}
\end{equation}
where \(^Br_{BT} = [r_x,r_y,r_z]^T\) denotes a vector from the origin of frame \{B\} to the origin of frame \{T\}, expressed in \{B\}. Based on the \(^BU_T\), the force/moment and velocity vectors can be transformed between frames as \cite{zhu2010virtual},

\begin{equation}\label{equ3}
^T\mathcal{V} =\, ^BU_T^T\,^B\mathcal{V},\quad ^B\mathcal{F} =\, ^BU_T\, ^T\mathcal{F}.
\end{equation}
Then, the dynamic equation of the free rigid body expressed in frame \{B\} can be derived as follows \cite{hejrati2022decentralized}:

\begin{equation}\label{equ4}
M_B\frac{d}{dt}(^B\mathcal{V})+C_B (^{B}\omega)\,{^B\mathcal{V}}+G_B=\, ^B\mathcal{F}^*
\end{equation}
where \(M_B \in \Re^{6\times6} \) is the mass matrix, \(C_B \in \Re^{6\times6} \) is the centrifugal and Coriolis matrix, \(G_B \in \Re^6 \) is the gravity vector, and \(^B\mathcal{F}^* \in \Re^6 \) is the net force/moment vector applied to the rigid body. A detailed formulation of the matrices is provided in \cite{hejrati2022decentralized}.

In the VDC approach, the required velocity is utilized instead of the desired velocity. The required velocity encompasses the desired velocity along with one or two error terms related to the position or force error in the position or force control mode, respectively. In this study, the required joint velocity is defined as,
\begin{equation}\label{equ5}
    \Dot{q}_r = \Dot{q}_d + \lambda(q_d - q)
\end{equation}
where \(\lambda\) is a positive constant, \(q\) is the joint angle, and \(\Dot{q}_d\) and \(q_d\) are the desired joint angular velocity and desired angle, respectively. The required linear/angular velocity vector \(^B\mathcal{V}_r\in \Re^6\) defined in frame \{B\} can be derived by utilizing (\ref{equ5}) and the kinematic relations from the base to the end-effector. Then, the required net force/moment vector can be defined as,
\begin{equation}\label{equ6}
^B\mathcal{F}_r^* = M_B\frac{d}{dt}(^B\mathcal{V}_r)+C_B(^{B}\omega)\,{^B\mathcal{V}_r}+G_B +\,^B\mathcal{F}_c
\end{equation}
where \(^B\mathcal{F}_c\) is the VDC control term. In the original VDC \cite{zhu2010virtual}, \(^B\mathcal{F}_c\) is only a proportional velocity term that engenders control performance problems in experiments and is difficult to tune the corresponding gain parameter. In this study, a much more sophisticated term is proposed to accomplish control objectives. Furthermore, an integral velocity term is added to \(^B\mathcal{F}_c\), which removes control performance problems.

The VDC approach breaks down the entire complex system into subsystems at each VCP and ensures the stability of the entire system by means of the virtual stability concept. VPFs will be deﬁned and used to characterize the dynamic interactions among subsystems. The introduction of this terminology is an important step leading to the deﬁnition of virtual stability.

\textbf{Deﬁnition 1.} \cite{zhu2010virtual} A non-negative accompanying function \(\nu(t) \in \Re\) is a piecewise, differentiable function possessing two properties: \(\nu(t) \geq 0\) for \(t > 0\), and \(\Dot{\nu}(t)\) exists almost everywhere.

\textbf{Deﬁnition 2.} \cite{zhu2010virtual} With respect to frame \{B\}, the VPF is deﬁned as the inner product of the linear/angular velocity vector error and the force/moment vector error, that is,

\begin{equation}\label{equ7}
p_B = (^B\mathcal{V}_r-^B\mathcal{V})^T(^B\mathcal{F}_r-\,^B\mathcal{F})
\end{equation}
where \(^B\mathcal{V}_r \in \Re^6\) and \(^B\mathcal{F}_r \in \Re^6\) represent the required vectors of \(^B\mathcal{V} \in \Re^6\) and \(^B\mathcal{F} \in \Re^6\), respectively.

\textbf{Deﬁnition 3.} \cite{zhu2010virtual} A subsystem that is virtually decomposed from a complex robot is said to be virtually stable with its affiliated vector \(\mathcal{X}(t)\) being a virtual function and its affiliated vector \(y(t)\) being a virtual function if and only if there exists a non-negative accompanying function
\begin{equation}\label{equ8}
\nu(t) \geq \frac{1}{2}\mathcal{X}(t)^TP\mathcal{X}(t)
\end{equation}
such that,
\begin{equation}\label{equ9}
\Dot{\nu}(t) \leq -y(t)^T\,Q\,y(t)+p_B-p_T
\end{equation}
holds, subject to,
\begin{equation}\label{equ10}
    \int_{0}^{\infty} s(t) dt\geq -\gamma_s
\end{equation}
with \(0\leq\gamma_s<\infty\), where \(P\) and \(Q\) are two block-diagonal positive-definite matrices, and \(p_B\) and \(p_T\) denote the sum of VPFs in the sense of Definition 2 at frames \{B\} (placed at driven VCPs) and \{T\} (placed at driving VCPs).

\textbf{Theorem 1.} \cite{zhu2010virtual} Consider a complex robot that is virtually decomposed into subsystems. If all the decomposed subsystems are virtually stable in the sense of Definition 1, then the entire system is stable.

Theorem 1 is the most important theorem in the VDC context. It establishes the equivalence between the virtual stability of every subsystem and the stability of the entire complex robot. Consequently, it allows us to concentrate on ensuring the virtual stability of every subsystem in lieu of the stability of the entire complex robot.

\subsection{lemmas and Assumptions}

In this part, all the lemmas and assumptions that are utilized in this paper are presented. 

\textbf{Assumption 1.}\, For an external disturbance \(F_d(t)\) and for human exogenous torque \(\tau_h(t)\), the following conditions are satisfied:
\begin{equation*}
    |F_d(t)|\leq \delta_1, \quad |\tau_h(t)|\leq \delta_2
\end{equation*}
with \(\delta_1 \geq 0,\, \delta_2 \geq 0\) being unknown constants.

\textbf{Assumption 2.}\, For the unknown robot model uncertainties \(\Delta_{Fr}\) and \(\Delta_{{\tau}r}\), along with human arm unmodeled dynamics \(\Delta_{Fh}\) and \(\Delta_{{\tau}h} \), we have,
\begin{equation*}
    |\Delta_{Fr}|\leq \delta_3, \quad |\Delta_{{\tau}r}|\leq \delta_4, \quad |\Delta_{Fh}|\leq \delta_5, \quad |\Delta_{{\tau}h}|\leq \delta_6
\end{equation*}
where \(\delta_3,\, \delta_4,\, \delta_5,\, \delta_6\) are positive unknown constants.

These assumptions are made to make sure that the applied unknown disturbances and uncertainties considered in the mathematical model are bounded.

\textbf{Lemma 1.} \cite{chen2010robust} RBFNNs can be utilized to estimate an unknown continuous function \(Z(\chi): \Re^m \to \Re\) with the approximation of,
\begin{equation*}
    Z(\chi) = \hat{W}^T\Psi(\chi) + \hat{\varepsilon}
\end{equation*}
where \(\chi = [\chi_1,\chi_2,...,\chi_m]^T \in \Re^m\) is the input vector of the neural networks, \(\hat{W}\) is the weight vector of the neural networks, \(\Psi(\chi)\) is the basis function of the RBFNNs, and \(\hat{\varepsilon}\) is the approximation error. The optimal weight vector \(W^*\) can be expressed by,
\begin{equation*}
    W^* = \arg \min_{\hat{W} \in\, \Xi_N}  \lbrace \sup_{\chi \in\, \Xi_T}|\hat{Z}(\chi|\hat{W})-Z(\chi)| \rbrace
\end{equation*}
where \(\Xi_N = \lbrace \hat{W}|\lVert \hat{W} \rVert \leq \kappa \rbrace\) is a valid set of vectors with \(\kappa\) being a design value, \(\Xi_T\) is an allowable set of the state vectors, and \(\hat{Z}(\chi|\hat{W}) = \hat{W}^T\,\Psi(\chi)\). 

\textbf{Lemma 2.} \cite{tee2009barrier} For any positive constant \(k_b\), let \(\mathfrak{Z} := \lbrace \mathfrak{z} \in \Re : -k_b<\mathfrak{z}<k_b\rbrace \subset \Re\) and \(\mathcal{N} := \Re^\mathfrak{l}\times\mathfrak{Z} \subset \Re^{\mathfrak{l}+1} \) be open sets. Consider the system,
\begin{equation*}
    \Dot{\eta} = \hbar(t,\eta)
\end{equation*}
where \(\eta := [\omega,\mathfrak{z}]^T \in \mathcal{N}\), and \(\hbar : \Re_+ \times \mathcal{N} \to \Re^{\mathfrak{l}+1}\) is piecewise continuous in \(t\) and locally Lipschitz in \(\mathfrak{z}\), uniformly in \(t\), on \(\Re_+ \times \mathcal{N}\). Suppose that there exist functions \(\mathfrak{U}: \Re^\mathfrak{l} \to \Re_+\) and \(\mathfrak{V}: \mathfrak{Z} \to \Re_+\), continuously differentiable and positive definite in their respective domains, such that,
\begin{equation*}
    \mathfrak{V}(\mathfrak{z}) \to \infty\quad  as\quad \mathfrak{z} \to -k_b \quad or\quad \mathfrak{z} \to k_b
\end{equation*}
\begin{equation*}
    \delta_7(\lVert\omega\rVert)\leq\,\mathfrak{U}(\omega)\,\leq\delta_8(\lVert\omega\rVert)
\end{equation*}
where \(\delta_7\) and \(\delta_8\) are class \(\mathcal{K}_\infty\) functions. Let \(\mathrm{V}(\eta) := \mathfrak{V}(\mathfrak{z}) + \mathfrak{U}(\omega)\), and \(\mathfrak{z}(0)\) belong to the set \(\mathfrak{z} \in (-k_b,k_b)\). If the following inequality holds:
\begin{equation*}
    \Dot{\mathrm{V}} = \frac{\partial\mathrm{V}}{\partial\eta}\hbar\leq0
\end{equation*}
then \(\mathfrak{z}(t)\) remains in the open set \(\mathfrak{z} \in (-k_b,k_b)\, \forall t \in [0,\infty)\).

\textbf{Remark 1.} In Lemma 2, the state space is split into \(\mathfrak{z}\) and \(\omega\), where \(\mathfrak{z}\) is the state to be constrained, and \(\omega\) is the free state. The constrained state \(\mathfrak{z}\) requires the barrier function \(\mathfrak{V}\) to prevent it from reaching the limits \(-k_b\) and \(k_b\) , while the free states may involve quadratic functions.

\textbf{Lemma 3.} \cite{lee2018natural}\cite{hejrati2022decentralized} For any inertial parameter vector of a rigid body \(\phi_B\) there is a one-to-one linear map \(f:\Re^{10} \rightarrow S(4)\), such that,
\begin{equation*}
    f(\phi_B)= \mathcal{L}_B = \begin{bmatrix}
        0.5tr(\Bar{I}_B).\textbf{1}-\Bar{I}_B & h_B \\
        h^T_B & m_B
        \end{bmatrix}
\end{equation*}
\begin{equation*}
    f^{-1}(\phi_B) = \phi_B(m_B,h_B,tr(\Sigma_B).\textbf{1}-\Sigma_B)
\end{equation*}
where \(m_B\),\(h_B\), and \(\Bar{I}_B\) are the mass, first mass moment, and rotational inertia matrix, respectively. \(\mathcal{L}_B \in S(4)\) is pseudo inertia matrix and \(\Sigma_B = 0.5tr(\Bar{I}_B)-\Bar{I}_B\).

\textbf{Lemma 4.} \cite{ren2020adaptive} If \(\Pi\) is considered as a signal exerted to the system and \(\pi\) is the desired control signal that would accomplish the control goal in the absence of input constraint, the issue of input saturation along with input deadzone can be transformed into
an equivalent input saturation by adding the deadzone inverse. Then the equivalent saturation can be expressed as,
\begin{multline*}
        \Pi = \varphi_b(sat_v((\varphi_b^{+}(\pi)))) = sat_{v-b}(\pi) \\= \begin{cases} 
      v-b & if \quad \pi\geq (v-b) \\
      \pi & if \quad (-v+b)<\pi< (v-b) \\
      -v+b & if \quad \pi\leq (-v+b) 
   \end{cases}
\end{multline*}
with \(sat_v(a)\) and \(\varphi_b(a)\) defined as,
\begin{equation*}
    sat_v(a) = \begin{cases}
    v & if \quad a \geq v \\
    a & if \quad -v < a < v \\
    -v & if \quad a\leq -v
    \end{cases}
\end{equation*}
\begin{equation*}
    \varphi_b(a) = \begin{cases}
    a-b & if \quad a \geq b \\
    0 & if \quad -b < a < b \\
    a+b & if \quad a\leq -b
    \end{cases}
\end{equation*}
where \(v\) and \(b\) are unknown
saturation and deadzone parameters. Also, \(\varphi_b^{+}\) is the right inverse of \(\varphi_b\) which satisfies \(\varphi_b.\varphi_b^{+}= I \).

\textbf{Remark 2.} The equivalent saturation constraint in Lemma 4 can be handled by common approaches like the anti-windup strategy. Consider the error between the desired control signal and the applied control signal as \(\Delta = \Pi-\pi\). In this study, RBFNNs are employed to approximate this \(\Delta\) that will be explained in the following.

\section{ Decentralization and augmentation of the Human-Robot model}

\begin{figure}[t]
      \centering
      \includegraphics[width = 2.5in]{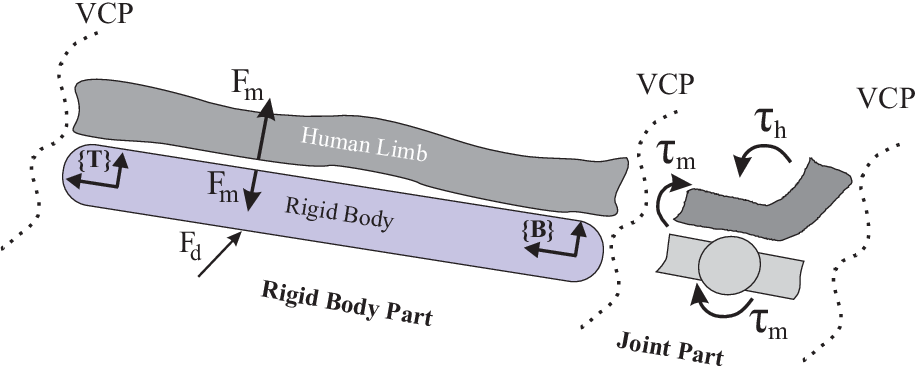}
      \caption{Human-robot augmented interaction model}
      \label{figurelabel1}
   \end{figure}

In this section, the original model in \cite{zhu2010virtual} and \cite{hejrati2022decentralized} is modified and a new decentralized and augmented pHRI model is proposed. As it is displayed in Fig. 2, the robot dynamics is merged with human arm dynamics through the VDC context, which enables us to design a decentralized controller to accomplish safe interaction between human and robot.

\subsection{Rigid Body Part}
Inspired by (\ref{equ4}) and Fig. 2, the dynamics of a rigid body in free space in the presence of unmodeled dynamics expressed in frame \{B\} can be explained as,
\begin{equation}\label{equ11}
\begin{split}
    M_B\frac{d}{dt}(^B\mathcal{V})+C_B(^{B}\omega)\,{^B\mathcal{V}}+ G_B + \Delta_{Fr} =\mathcal{B}_1(\,^B\mathcal{F}^*)\\
    -\Bar{F}_d - \Bar{F}_m
\end{split}
\end{equation}
where \(\mathcal{B}_1(.)\) is the internal equivalent saturation function in the sense of Lemma 4, and \(\Bar{F}_d = \, ^dU_B\, ^BF_d\) and \(\Bar{F}_m = \, ^mU_B\, ^BF_m\) with \(F_d\) and \(F_m\) being the unknown disturbance and interaction force between human arm link and robot rigid body, respectively, with \(^dU_B\) and \(^mU_B\) being unknown transformation matrix. Also, \(\Delta_{Fr}\) is the unknown unmodeled dynamics of the rigid body. Utilizing the VCP concept, dynamics of the human arm's link expressed in frame \{B\}, can be written as,
\begin{equation}\label{equ12}
\begin{split}
    ^hM_{B}\frac{d}{dt}(^{B}\mathcal{V}_h)+\,^hC_{B}\,^{B}\mathcal{V}_h+\,^hG_{B} + \Delta_{Fh} =\, F_m
\end{split}
\end{equation}
where \(\Delta_{Fh}\) is the unknown unmodeled dynamics of human arm link and \(^h(.)\) and \((.)_h\) stands for human. Therefore, by substituting (\ref{equ12}) in (\ref{equ11}) and using Lemma 4 and remark
2 as \(\mathcal{B}_1(\,^B\mathcal{F}^*) = \,^B\mathcal{F}^*-\Delta\mathcal{F}^*\) along with defining \(\Delta_D = \Delta_{Fh} + \Delta_{Fr}+\Delta\mathcal{F}^*\), we obtain,
\begin{equation}\label{equ13}
\begin{split}
    \mathcal{M}_B\frac{d}{dt}(^B\mathcal{V})+\,\mathcal{C}_B(^{B}\omega)\,{^B\mathcal{V}}+\,\mathcal{G}_B =\,^B\mathcal{F}^*- F_d -\Delta_D 
\end{split}
\end{equation}
where \(\mathcal{M}_B =\, ^hM_{B}+ M_B\), \(\mathcal{C}_B =\, ^hC_{B}+ C_B\), and \(\mathcal{G}_B =\, ^hG_{B}+ G_B\).

\subsection{Actuator Part}

The actuator dynamics of the robot (Fig. 2) in the presence of unknown uncertainty and input constraints can be expressed as,
\begin{equation}\label{equ14}
    I_m \Ddot{q} + \Delta_{{\tau}r} = \mathcal{B}_2(\tau^*)-\tau_m
\end{equation}
where \(\mathcal{B}_2(.)\) is the internal equivalent saturation function in the sense of Lemma 4, with \(I_m\) being motor inertia, \(\tau^*=\tau_i-\tau_{ai}\) being net control torque, where \(\tau_i\) is control signal and \(\tau_{ai}\) being defined later, and \(\tau_m\) being interaction torque that drives human joint. \(\Delta_{{\tau}r}\) is the unknown actuator uncertainty, which is estimated by employing RBFNNs. In actuators with a complex power transmission mechanism, such an estimation can increase the performance of the controller. Similar to (\ref{equ14}), the human joint dynamics can be written as,
\begin{equation}\label{equ15}
    I_h \Ddot{q} + \Delta_{{\tau}h} = \tau_m - \tau_h
\end{equation}
where \(I_h\) is the equivalent inertia of the human joint and \(\tau_h\) is HET around the joint created by muscles. Substituting (\ref{equ15}) into (\ref{equ14}) and utilizing Lemma 4 and remark 2 as \(\mathcal{B}_2(\tau^*) = \tau^*-\Delta\tau^*\) along with defining \(\Delta_J = \Delta_{{\tau}h} + \Delta\tau^* + \Delta_{{\tau}r}\), we obtain,
\begin{equation}\label{equ16}
    \mathcal{I} \Ddot{q} = \tau^* - \tau_h - \Delta_J
\end{equation}
where \(\mathcal{I} = I_m + I_h\). 

The new equations specified in (\ref{equ13}) and (\ref{equ16}) are modified representations of rigid body and joint dynamics of the robot augmented with human arm dynamics in a decentralized way. In these equations, input constraint and unmodeled dynamics in the actuator part and rigid body, as well as in the human arm model, are considered. In the following section, a controller design procedure is explained for the derived pHRI model.

\section{Control Design and Stability Proof}

In this section, a decentralized controller is designed to stabilize the pHRI and accomplish the control objectives of the system. Fig. 3 displays the details of the VDC decomposition by means of VCP.

\begin{figure}[t]
      \centering
      \subfloat[]{\includegraphics[width = 0.4\textwidth]{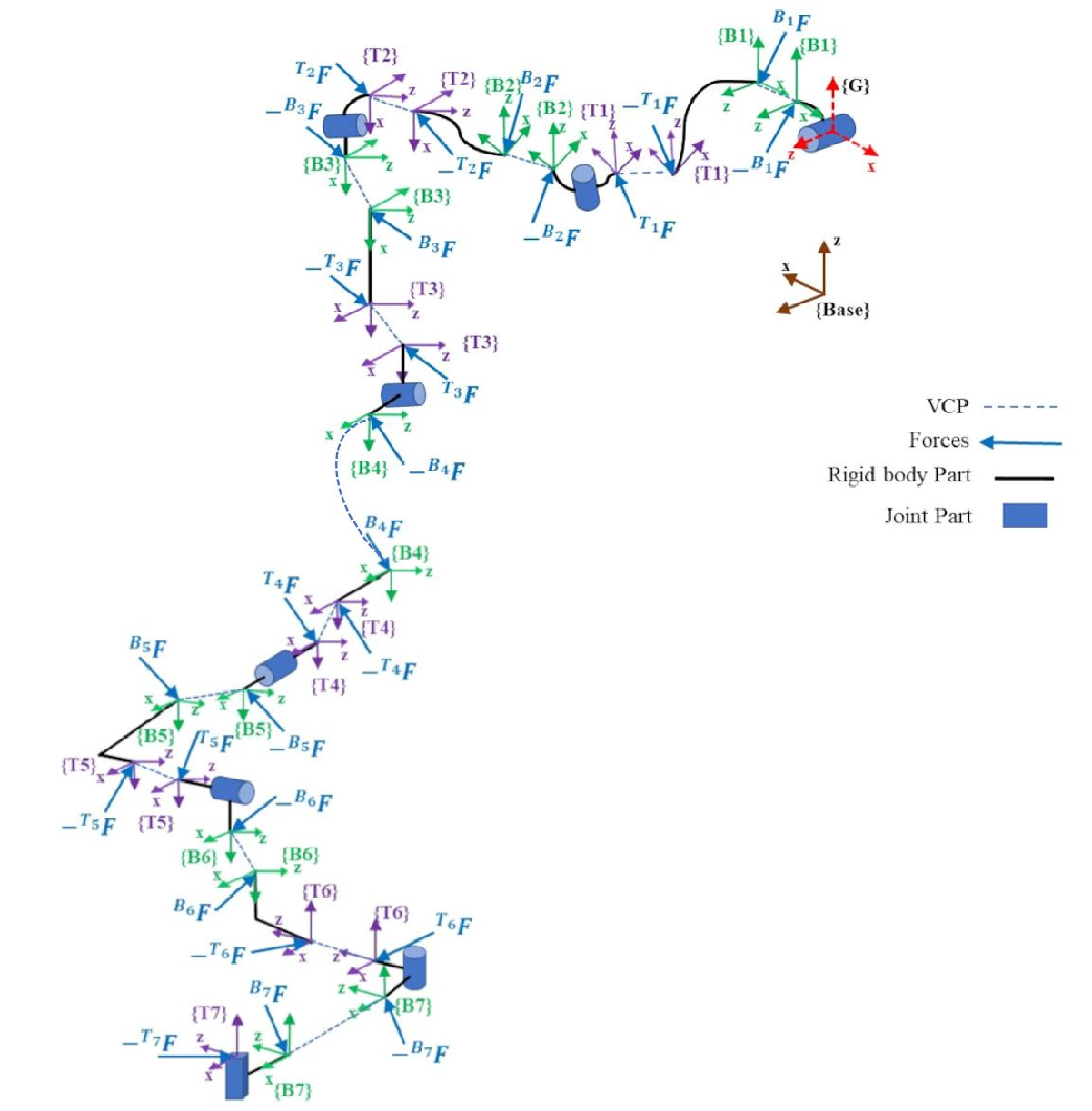}
      \centering
      \label{fig4}}
      \hfil
      \subfloat[]{\includegraphics[width = 0.4\textwidth]{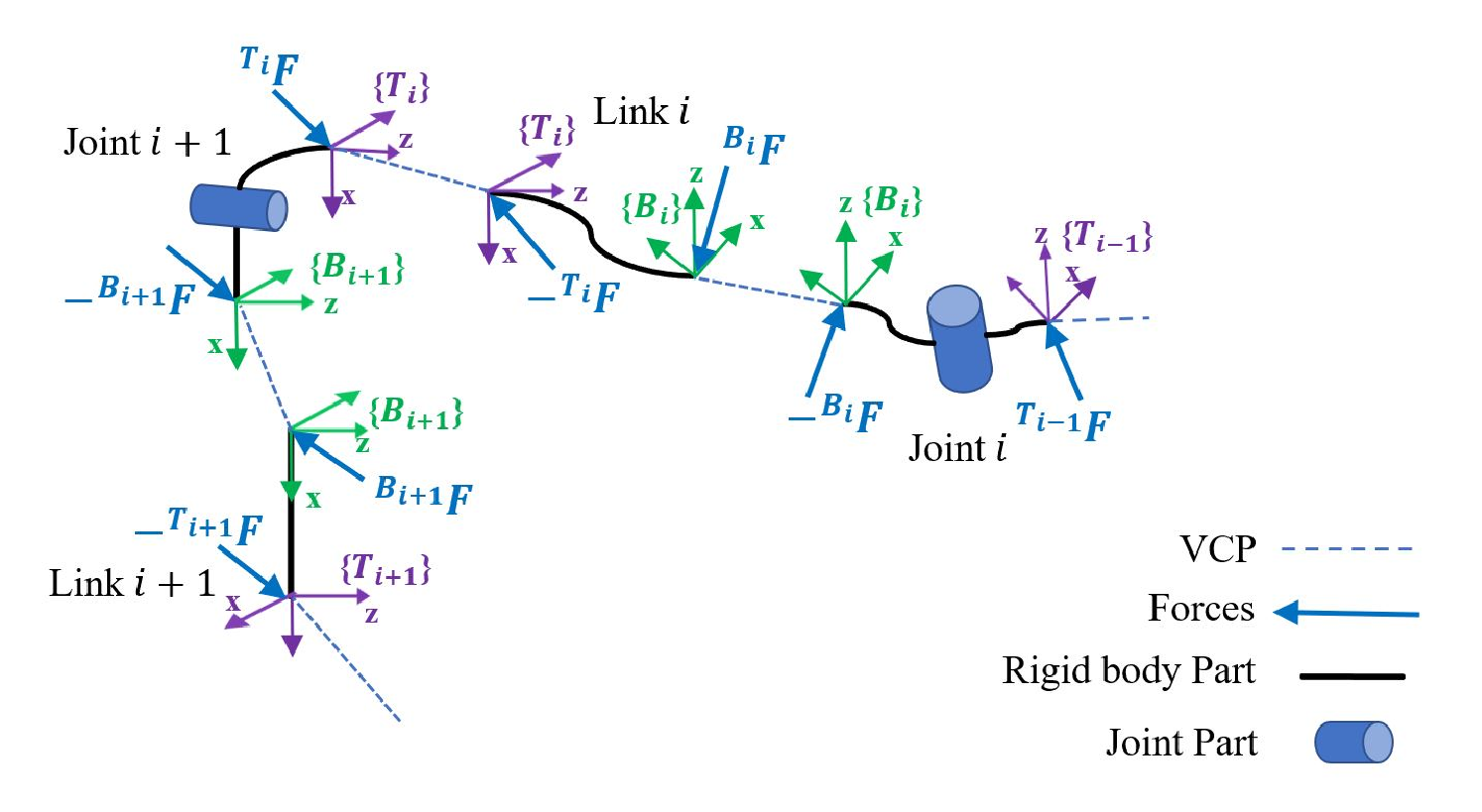}
      \centering
      \label{fig5}}
      \caption{Decomposition of pHRI model for decentralized control design. a) virtually decomposed form of entire system with attached VDC frames and VCP, b) ith subsystem of decomposed model}
   \end{figure}

\subsection{Force/Velocity Computation}

The linear/angular velocity vectors can be computed \cite{zhu2010virtual} using kinematic relationships as according to Fig. 3,
\begin{equation}\label{equ17}
     ^{B_i}\mathcal{V} =\, ^{T_{i-1}}U_{B_i}^T\, ^{T_{i-1}}\mathcal{V} + \mu_i \Dot{q}_i 
\end{equation}
\begin{equation}\label{equ18}
     ^{T_i}\mathcal{V} =\, ^{B_i}U_{T_i}^T\, ^{B_i}\mathcal{V}
\end{equation}
where \(i = 1 ... 7\), \(^{T_0}\mathcal{V}=\, ^G\mathcal{V}\) and \(^{T_7}\mathcal{V}\) are the linear/angular velocity vector of the ground and end-effector, respectively, \(q_i\) and \(\Dot{q}_i\) are the joint angles and velocities, respectively, \(\mu_i = z_\tau\) for \(i = 1,2,3,4,6\), \(\mu_i = y_\tau\) for \(i = 7\), and \(\mu_i = x_\tau\) for \(i = 5\), which \(z_\tau = [0,0,0,0,0,1]^T\), \(y_\tau = [0,0,0,0,1,0]^T\), and \(x_\tau = [0,0,0,1,0,0]^T\). Moreover, \(^{B_i}U_{T_i}\) can be computed with (\ref{equ1}) by replacing \(B\) and \(T\) with \(B_i\) and \(T_i\), respectively. The required linear/angular velocities can also be expressed as
\begin{equation}\label{equ19}
     ^{B_i}\mathcal{V}_r =\, ^{T_{i-1}}U_{B_i}^T\, ^{T_{i-1}}\mathcal{V}_r + \mu_i \Dot{q}_{ri}
\end{equation}
\begin{equation}\label{equ20}
     ^{T_i}\mathcal{V}_r =\, ^{B_i}U_{T_i}^T\, ^{B_i}\mathcal{V}_r
\end{equation}
where  \(^{T_0}\mathcal{V}_r =\, ^G\mathcal{V}_r\) and \(^{T_7}\mathcal{V}_r\) are the required linear/angular velocity vectors of the ground and end-effector, respectively, and \(\Dot{q}_{ri}\) is the required joint angular velocity defined as,
\begin{equation}\label{equ21}
    \Dot{q}_{ri} = \Dot{q}_{di} + \lambda_i(q_{di}-q_i)
\end{equation}
where \(\lambda_i\) are positive constants. The force/moment vectors can be calculated as,
\begin{equation}\label{equ22}
    ^{B_j}\mathcal{F} =\, ^{B_j}U_{T_j}\, ^{T_j}\mathcal{F} +\, ^{B_j}\mathcal{F}^*
\end{equation}
\begin{equation}\label{equ23}
    ^{T_{j-1}}\mathcal{F} =\, ^{T_{j-1}}U_{B_j}\, ^{B_j}\mathcal{F} 
\end{equation}
with required terms as,
\begin{equation}\label{equ24}
    ^{B_j}\mathcal{F}_r =\, ^{B_j}U_{T_j}\, ^{T_j}\mathcal{F}_r +\, ^{B_j}\mathcal{F}_r^*
\end{equation}
\begin{equation}\label{equ25}
    ^{T_{j-1}}\mathcal{F}_r =\, ^{T_{j-1}}U_{B_j}\, ^{B_j}\mathcal{F}_r
\end{equation}
where \(^{T_7}\mathcal{F}\) is the interaction force between the end-effector and environment, and \(^{T_0}\mathcal{F}\) is the applied force from joint 1 to the ground for \(j = 7 ... 1\). \(^{B_j}F_r^*\) is the required net force/moment vector defined in (\ref{equ30}). 

The required net torques and forces for joints and rigid bodies, respectively, are actions that make the corresponding subsystems behave in a desired way, establishing the control goals of the overall system. The unified control command that will be transmitted to the motor to accomplish the task can be computed as below,
\begin{equation}\label{equ26}
    \tau_i = \tau^*_{ri} + \tau_{ari}
\end{equation}
where \(\tau^*_{ri}\) is defined in (\ref{equ41}) and \(\tau_{ari}\) is a term related to rigid body dynamics, computed as,
\begin{equation}\label{equ27}
    \tau_{ari} = \mu_i^T\, ^{B_i}\mathcal{F}_r.
\end{equation}
with \(i = 1...7\) and \(^{B_i}\mathcal{F}_r\) defined in (\ref{equ24}).

In this section, the force/velocity and corresponding required terms are computed, which are important to compute the control action defined in (\ref{equ26}). A detailed explanation of the control design (\(\tau^*_{ri}\) and \(^{B_i}\mathcal{F}^*\)) and stability analysis corresponding to each subsystem is provided in the following.

\mbox{\textbf{Remark 3.}} The local control law in \mbox{(\ref{equ26})}, has two terms to consider both dynamics; the first term computes the required action to achieve the control goal of the subsystem, while the second term compensates for the dynamics of the rigid body such as gravitational and inertia effects. This independent design approach for each term results in the avoidance of nonlinearity in the coupled rigid body and actuator dynamics. However, the coupling nonlinearity is taken into account in the overall control law \mbox{(\ref{equ26})}.

\subsection{Rigid Body Part}

As mentioned earlier, \(\Delta_D\) in (\ref{equ13}) is an unknown unmodeled dynamics that can be estimated by RBFNNs using Lemma 1,
\begin{equation}\label{equ28}
    \Delta_{Di} = W^{*T}_{Di}\Psi(\chi_{Di})+\varepsilon^*_{Di}
\end{equation}
where \(i = 1...,7\), \(W^*_{Di} \in \Re^{\mathbf{j}\times6}\) is the optimal weights, \(\chi_{Di}\) is the input vector \(\chi_{Di} = [\, ^{B_i}\mathcal{V}_r,\, ^{B_i}\mathcal{V},\, ^{B_i}E_I, ^{B_i}E_D,\tau_{ari}]^T\) with \(^{B_i}E_I = \int_{0}^{t} (\,^{B_i}\mathcal{V}_r-\,^{B_i}\mathcal{V}) dt\), \(^{B_i}E_D =\, ^{B_i}\mathcal{V}_r - ^{B_i}\mathcal{V}\), and \(\varepsilon^*_{Di} = \Bar{\varepsilon}^*_{Di}+F_{di}\) where \(\Bar{\varepsilon}^*_{Di} \in \Re^{6}\) is the error vector of RBFNNs, and \(\mathbf{j}\) is the number of neuron units in RBFNNs. Then, replacing \(B\) with \(B_i\) in (\ref{equ13}) and utilizing the linear-in-parameter feature, one can write,
\begin{equation}\label{equ29}
\begin{split}
    \mathcal{M}_{B_i}\frac{d}{dt}(^{B_i}\mathcal{V}_r)+\,\mathcal{C}_{B_i}(^{B_i}\omega)\,{^{B_i}\mathcal{V}_r}+\,\mathcal{G}_{B_i} =\,Y_{B_i}\phi_{B_i} 
\end{split}
\end{equation}
where \(i = 1...,7\), \(Y_{B_i} \in \Re^{6\times10}\) is the compact regressor matrix defined in \cite{hejrati2022decentralized} and \(\phi_{B_i} \in \Re^{10}\) are the constant inertial parameters. In the following theorem, the stability of the rigid body subsystems is ensured.

\textbf{Theorem 2.} For the decentralized model of pHRI with a rigid body part represented by the dynamic model of (\ref{equ13}), the corresponding decentralized control action can be designed as,
\begin{equation}\label{equ30}
\begin{split}
    ^{B_i}\mathcal{F}^*_r &=\,\mathcal{K}_{Di}(\,^{B_i}\mathcal{V}_r-\,^{B_i}\mathcal{V}) + \hat{W}_{Di}^T\Psi(\chi_{Di})+\hat{\varepsilon}_{Di} \\
    &+\mathcal{K}_{Ii} \int_{0}^{t} (\,^{B_i}\mathcal{V}_r-\,^{B_i}\mathcal{V}) dt +Y_{B_i}\hat{\phi}_{B_i}
\end{split}
\end{equation}
with the update rules as below,
\begin{equation}\label{equ31}
\begin{split}
    \Dot{\hat{\mathcal{L}}}_{B_i} = \frac{1}{\gamma_1}\hat{\mathcal{L}}_{B_i}\,\mathcal{S}_{B_i}\,\hat{\mathcal{L}}_{B_i}
\end{split}
\end{equation}
\begin{equation}\label{equ32}
\begin{split}
    \Dot{\hat{W}}_{Di} = \Gamma_{B_i}\,\Psi(\chi_{Di})\,(\,^{B_i}\mathcal{V}_r-\,^{B_i}\mathcal{V})
\end{split}
\end{equation}
\begin{equation}\label{equ33}
\begin{split}
    \Dot{\hat{\varepsilon}}_{Di} = \gamma_{2i}\,(\,^{B_i}\mathcal{V}_r-\,^{B_i}\mathcal{V})
\end{split}
\end{equation}
such that the virtual stability of rigid body part is ensured in the sense of Definition 3. The \(\hat{(.)}\) denotes estimated value, \(\gamma_1\) and \(\gamma_{2i}\) are positive constants, and \(\Gamma_{B_i}\) are positive-definite constants. \(\hat{\mathcal{L}}_{B_i}\) is derived from \(\hat{\phi}_{B_i}\) utilizing Lemma 3.

\textbf{Proof.} Replacing \(B\) with \(B_i\) in (\ref{equ13}) and subtracting it from (\ref{equ30}) and utilizing (\ref{equ28}) and (\ref{equ29}), we have,
\begin{equation}\label{equ34}
\begin{split}
    &^{B_i}\mathcal{F}^*_r-\,^{B_i}\mathcal{F}^*=\,\mathcal{K}_{Di}(\,^{B_i}\mathcal{V}_r-\,^{B_i}\mathcal{V}) + Y_{B_i}\Tilde{\phi}_{B_i}\\
    &+\mathcal{K}_{Ii} \int_{0}^{t} (\,^{B_i}\mathcal{V}_r-\,^{B_i}\mathcal{V}) dt + \Tilde{W}_{Di}^T\Psi(\chi_{Di})+\Tilde{\varepsilon}_{Di} \\
    &+ \mathcal{M}_{B_i}\frac{d}{dt}(^{B_i}\mathcal{V}_r-\,^{B_i}\mathcal{V})+\,\mathcal{C}_{B_i}(^{B_i}\mathcal{V}_r-\,^{B_i}\mathcal{V})
\end{split}
\end{equation}
where \(\Tilde{(.)}=\hat{(.)}-(.)^*\) for \(\Tilde{W}\) and \(\Tilde{\varepsilon}\), and \(\Tilde{(.)}=\hat{(.)}-(.)\) for \(\Tilde{\phi}\).

Based on Definition 1, the non-negative accompanying function candidate for stability analysis is,
\begin{equation}\label{equ35}
\begin{split}
    \nu_i(t) &=\, \frac{1}{2} (^{B_i}\mathcal{V}_r-\,^{B_i}\mathcal{V})^T\, \mathcal{M}_{B_i}\,(^{B_i}\mathcal{V}_r-\,^{B_i}\mathcal{V})\\ 
    &+ \frac{1}{2} \lbrace\int_{0}^{t} (\,^{B_i}\mathcal{V}_r-\,^{B_i}\mathcal{V}) dt\rbrace^T\,\mathcal{K}_{Ii} \lbrace\int_{0}^{t} (\,^{B_i}\mathcal{V}_r-\,^{B_i}\mathcal{V}) dt\rbrace\\ 
    &+\gamma_1\mathcal{D}_F(\mathcal{L}_{B_i}\rVert \hat{\mathcal{L}}_{B_i}) + \frac{1}{2}tr(\Tilde{W}_{Di}^T\,\Gamma_{B_i}^{-1}\Tilde{W}_{Di}) 
    \\
    &+  \frac{1}{2\gamma_{2i}} \Tilde{\varepsilon}_{Di}^T\,\Tilde{\varepsilon}_{Di}.
\end{split}
\end{equation}
where \(\mathcal{D}_F(\mathcal{L}_{B_i}\rVert \hat{\mathcal{L}}_{B_i}) \) defined as \cite{hejrati2022decentralized},
\begin{equation}\label{equ36}
    \mathcal{D}_F(\mathcal{L}_{B_i}\rVert \hat{\mathcal{L}}_{B_i}) = log\frac{|\hat{\mathcal{L}}_{B_i}|}{|\mathcal{L}_{B_i}|}+tr(\hat{\mathcal{L}}_{B_i}^{-1}\mathcal{L}_{B_i})-4
\end{equation}
is the Bregman divergence metric. Taking the derivative of (\ref{equ35}) and using (\ref{equ34}), leads to,
\begin{equation}\label{equ37}
    \begin{split}
        \Dot{\nu}_i(t) &= (^{B_i}\mathcal{V}_r-\,^{B_i}\mathcal{V})^T(\,^{B_i}\mathcal{F}^*_r -\,^{B_i}\mathcal{F}^*)
        \\
        &+ \gamma_1 tr([\hat{\mathcal{L}}_{B_i}^{-1}\Dot{\hat{\mathcal{L}}}_{B_i}
        \hat{\mathcal{L}}_{B_i}^{-1}]\widetilde{\mathcal{L}}_{B_i})-(^{B_i}\mathcal{V}_r-\,^{B_i}\mathcal{V})^T\,Y_{B_i}\Tilde{\phi}_{B_i}\\
        &-(^{B_i}\mathcal{V}_r-\,^{B_i}\mathcal{V})^T\,\mathcal{K}_{Di}(\,^{B_i}\mathcal{V}_r-\,^{B_i}\mathcal{V})\\
        &-(^{B_i}\mathcal{V}_r-\,^{B_i}\mathcal{V})^T\,\Tilde{W}_{Di}^T\Psi(\chi_{Di})\\
        &-(^{B_i}\mathcal{V}_r-\,^{B_i}\mathcal{V})^T\,\Tilde{\varepsilon}_{Di}\\
        &-(^{B_i}\mathcal{V}_r-\,^{B_i}\mathcal{V})^T\,\mathcal{C}_{B_i}(^{B_i}\mathcal{V}_r-\,^{B_i}\mathcal{V}) \\
        &+\frac{1}{\gamma_{2i}} \Tilde{\varepsilon}_{Di}^T \Dot{\Tilde{\varepsilon}}_{Di}+ tr(\Dot{\Tilde{W}}_{Di}^T\,\Gamma_{B_i}^{-1}\Tilde{W}_{Di}).
    \end{split}
\end{equation}
Recalling Lemma 3, defining \(s_{B_i} = Y_{B_i}^T\,(^{B_i}\mathcal{V}_r-\,^{B_i}\mathcal{V})\), exploiting the relationship below,
\begin{equation*}
    (^{B_i}\mathcal{V}_r-\,^{B_i}\mathcal{V})^T\,\Tilde{W}_{Di}^T\Psi(\chi_{Di}) = tr((^{B_i}\mathcal{V}_r-\,^{B_i}\mathcal{V})\,\Psi^T(\chi_{Di})\,\Tilde{W}_{Di})
\end{equation*}
and using \(\widetilde{\phi}_{B_i}^Ts_{Bi} = tr(\widetilde{\mathcal{L}}_{B_i} \mathcal{S}_{B_i})\), where \(\mathcal{S}_{B_i}\) is a unique symmetric matrix defined in Appendix A, along with the fact that \(\mathcal{C}_{Bi}\) is a skew-symmetric matrix, we can rewrite (\ref{equ37}) as,
\begin{equation}\label{equ38}
\begin{split}
    \Dot{\nu}_i(t) &= (^{B_i}\mathcal{V}_r-\,^{B_i}\mathcal{V})^T(\,^{B_i}\mathcal{F}^*_r -\,^{B_i}\mathcal{F}^*) \\
        &-\,tr(\lbrace \mathcal{S}_{B_i}-  \gamma_1[\hat{\mathcal{L}}_{B_i}^{-1}\Dot{\hat{\mathcal{L}}}_{B_i}\hat{\mathcal{L}}_{B_i}^{-1}]\rbrace\,\widetilde{\mathcal{L}}_{B_i}) \\
        &-(^{B_i}\mathcal{V}_r-\,^{B_i}\mathcal{V})^T\,\mathcal{K}_{Di}(\,^{B_i}\mathcal{V}_r-\,^{B_i}\mathcal{V})\\
        &-tr(\lbrace(^{B_i}\mathcal{V}_r-\,^{B_i}\mathcal{V})\,\Psi^T(\chi_{Di})-\Dot{\Tilde{W}}_{Di}^T\,\Gamma_{B_i}^{-1}\rbrace\Tilde{W}_{Di})\\
        &-\Tilde{\varepsilon}^T_{Di}((^{B_i}\mathcal{V}_r-\,^{B_i}\mathcal{V})\,-\frac{1}{\gamma_{2i}}\, \Dot{\Tilde{\varepsilon}}_{Di}).
\end{split}
\end{equation}
Finally, by substituting (\ref{equ31})-(\ref{equ33}) into (\ref{equ38}), one can get,
\begin{equation}\label{equ39}
    \begin{split}
        \Dot{\nu}_i(t) &= (^{B_i}\mathcal{V}_r-\,^{B_i}\mathcal{V})^T(\,^{B_i}\mathcal{F}^*_r -\,^{B_i}\mathcal{F}^*)  \\
        &-(^{B_i}\mathcal{V}_r-\,^{B_i}\mathcal{V})^T\,\mathcal{K}_{Di}(\,^{B_i}\mathcal{V}_r-\,^{B_i}\mathcal{V}).\qquad \blacksquare
    \end{split}
\end{equation}

\subsection{Actuator Dynamics Part}

In this section, a control signal is designed to stabilize joint dynamics and obtain control goals. Since the forces in the muscles generate torque around the joint, the estimation of the HET is considered in the actuator dynamics part. Utilizing Lemma 1, \(\Delta_J\) can be estimated as,
\begin{equation}\label{equ40}
    \Delta_{Ji} = W^{*T}_{Ji}\Psi(\chi_{Ji})+\varepsilon^*_{Ji}.
\end{equation}
where  \(i = 1...,7\), \(W^*_{Ji} \in \Re^{\mathbf{j}}\) is the optimal weights, \(\chi_{Ji}\) is the input vector \(\chi_{Ji} = [\Ddot{q}_{ri}, \Dot{q}_{ri},\Dot{q}_i, e_{ai},\Dot{e}_{ai},\tau^*_{ri}]^T\), and \(\varepsilon^*_{Ji} = \Bar{\varepsilon}^*_{Ji}+\tau_{hi}\) where \(\Bar{\varepsilon}^*_{Ji} \in \Re\) is the error vector of RBFNNs and \(\tau_{hi}\) is human exogenous torque represented at the \(i^{th}\) subsystem, where \(\mathbf{j}\) is the number of neuron units in RBFNNs. In the next theorem, the control design and stability analysis of joint dynamics is examined.

In this study, state constraint is considered and BLF is employed to ensure that the joint angles will not violate constraints. Joint angles, \(\mathbf{q} = [q_1,...,q_7]^T\), are required to remain in the set \(|\mathbf{q}|<k_c\) for \(t>0\), with \(k_c = [k_{c1},...,k_{c7}]^T\) being the positive constant standing for constraints. Moreover, the required angles, \(\mathbf{q}_r = [q_{1r},...,q_{7r}]^T\), and their higher derivatives, \(\Dot{\mathbf{q}}_r\) and \(\Ddot{\mathbf{q}}_r\), have a bound of \(|\mathbf{q}_r| < k_{cr}\), \(|\Dot{\mathbf{q}}_r| < k_{cr1} \), and \(|\Ddot{\mathbf{q}}_r| < k_{cr2} \). By integrating (\ref{equ21}) we have \(k_{cr} = k_{cd}+c\) with \(|\mathbf{q}_d|<k_{cd}\) and \(c\) being positive constant as the upper limit of tracking error integrator. Therefore, the constraint on error, \(\mathbf{e}_a = \mathbf{q}_r - \mathbf{q}\), can be derived as \(\mathbf{k}_b = k_{cr}-k_c\). If \(|\mathbf{e}_a|<\mathbf{k}_b\), then all joint angles are kept in the safe region without constraint violation as long as \(|\mathbf{e}(0)_a|<\mathbf{k}_b\). Theorem 3 summarizes the control law for the actuator dynamics part with state constraint.

\textbf{Theorem 3.} For the decentralized model of pHRI with actuator dynamics represented by the dynamic model of (\ref{equ16}), the corresponding decentralized control action can be designed as,
\begin{equation}\label{equ41}
\begin{split}
    \tau^*_{ri} &= k_{di}(\Dot{q}_{ri}-\,\Dot{q}_i) + k_{Ii} \int_{0}^{t} (\Dot{q}_{ri}-\,\Dot{q}_i) dt\\ &+Y_{ai}\hat{\phi}_{ai} + \hat{W}_{Ji}^T\Psi(\chi_{Ji})+\hat{\varepsilon}_{Ji}+\frac{e_{ai}}{k^2_{bi}-e^2_{ai}}
\end{split}
\end{equation}
with the update rules as follows:
\begin{equation}\label{equ42}
\begin{split}
    \Dot{\hat{\mathcal{L}}}_{ai} = \frac{1}{\zeta}\hat{\mathcal{L}}_{ai}\,\mathcal{S}_{ai}\,\hat{\mathcal{L}}_{ai}
\end{split}
\end{equation}
\begin{equation}\label{equ43}
\begin{split}
    \Dot{\hat{W}}_{Ji} = \beta_{1i}\,(\Dot{q}_{ri}-\,\Dot{q}_i)\,\Psi(\chi_{Ji})
\end{split}
\end{equation}
\begin{equation}\label{equ44}
\begin{split}
    \Dot{\hat{\varepsilon}}_{Ji} = \beta_{2i}\,(\Dot{q}_{ri}-\,\Dot{q}_i)
\end{split}
\end{equation}
such that the virtual stability of the actuator dynamics part is ensured in the sense of Definition 3. The \(\hat{(.)}\) denotes the estimated value, \(\zeta\), \(\beta_{1i}\), and \(\beta_{2i}\) are positive constants. \(\hat{\mathcal{L}}_{ai}\) is derived from \(\hat{\phi}_{ai}\) utilizing Lemma 3. Moreover, \(e_a = q_r-q\) and \(k_{bi}\) are in the sense of Lemma 2.

\textbf{Proof.} Subtracting (\ref{equ16}) from (\ref{equ41}) and using (\ref{equ40}), we have,
\begin{equation}\label{equ45}
    \begin{split}
        \tau^*_{ri}-\,\tau^*_i &=\,k_{di}(\Dot{q}_{ri}-\,\Dot{q}_i) + k_{Ii} \int_{0}^{t} (\Dot{q}_{ri}-\,\Dot{q}_i) dt +Y_{ai}\Tilde{\phi}_{ai} \\ &+\Tilde{W}_{Ji}^T\Psi(\chi_{Ji})+\Tilde{\varepsilon}_{Ji} + \mathcal{I}\,\frac{d}{dt}(\Dot{q}_{ri}-\,\Dot{q}_i)+\frac{e_{ai}}{k^2_{bi}-e^2_{ai}}.
    \end{split}
\end{equation}
Then, the accompanying function corresponding to the actuator part can be defined as,
\begin{equation}\label{equ46}
    \begin{split}
        \nu_{ai}(t) &=\, \frac{1}{2}\mathcal{I} (\Dot{q}_{ri}-\,\Dot{q}_i) + \frac{1}{2}k_{Ii} \lbrace\int_{0}^{t} (\Dot{q}_{ri}-\,\Dot{q}_i) dt\rbrace^2\\
        &+\zeta\mathcal{D}_F(\mathcal{L}_{ai}\rVert \hat{\mathcal{L}}_{ai}) + \frac{1}{2\beta_{2i}} \Tilde{\varepsilon}_{Ji}^2 + \frac{1}{2\beta_{1i}}\Tilde{W}_{Ji}^T\,\Tilde{W}_{ji}\\
        &+\frac{1}{2}\log{\frac{k^2_{bi}}{k^2_{bi}-e^2_{ai}}}.
    \end{split}
\end{equation}
where \(\mathcal{D}_F(\mathcal{L}_{ai}\rVert \hat{\mathcal{L}}_{ai}) \) is defined the same as (\ref{equ36}). Taking the derivative of (\ref{equ46}) and using (\ref{equ45}), one can obtain,
\begin{equation}\label{equ47}
    \begin{split}
        \Dot{\nu}_{ai}(t) &= (\Dot{q}_{ri}-\,\Dot{q}_i)(\tau^*_{ri}-\,\tau^*_i) + \zeta tr([\hat{\mathcal{L}}_{ai}^{-1}\Dot{\hat{\mathcal{L}}}_{ai}\hat{\mathcal{L}}_{ai}^{-1}]\widetilde{\mathcal{L}}_{ai}) \\
        &-k_{di}(\Dot{q}_{ri}-\,\Dot{q}_i)^2
        -(\Dot{q}_{ri}-\,\Dot{q}_i)\,Y_{ai}\Tilde{\phi}_{ai}+\frac{\Dot{e}_{ai}e_{ai}}{k^2_{bi}-e^2_{ai}} \\
        &-(\Dot{q}_{ri}-\,\Dot{q}_i)\,\Tilde{W}_{Ji}^T\Psi(\chi_{Ji})
        -(\Dot{q}_{ri}-\,\Dot{q}_i)\,\Tilde{\varepsilon}_{Ji}\\
        &-(\Dot{q}_{ri}-\,\Dot{q}_i)\,\frac{e_{ai}}{k^2_{bi}-e^2_{ai}} +\frac{1}{\beta_{2i}} \Tilde{\varepsilon}_{Ji} \Dot{\Tilde{\varepsilon}}_{Ji}+\frac{1}{\beta_{1i}} \Dot{\Tilde{W}}_{Ji}^T\,\Tilde{W}_{Ji}.
    \end{split}
\end{equation}
Now, by defining \(s_{ai} = Y^T_{ai}(\Dot{q}_{ri}-\Dot{q}_i)\), replacing (\ref{equ42})-(\ref{equ44}) in (\ref{equ47}), and using the fact that \(\Dot{e}_a=\Dot{q}_r-\Dot{q}\), we get to,
\begin{equation}\label{equ48}
    \begin{split}
        \Dot{\nu}_{ai}(t) = (\Dot{q}_{ri}-\,\Dot{q}_i)(\tau^*_{ri}-\,\tau^*_i)-k_{di}(\Dot{q}_{ri}-\,\Dot{q}_i)^2. \quad \blacksquare
    \end{split}
\end{equation}

In the control signal (\ref{equ41}), the last term is related to the state constraint, which is derived from the logarithmic Lyapunov function. In this constraint term, defining error based on the required angle applies much stricter control action to avoid violating the constraints, resulting in a very small tracking error. Moreover, human exogenous torque \(\tau_{hi}\) is estimated along with the error vector of RBFNNs.

In the following, the final theorem for stability analysis of the entire system is provided.

\textbf{Theorem 4.} The entire pHRI model, which is represented by (\ref{equ13}) and (\ref{equ16}) in the decentralized model, is asymptotically stable in the sense of Theorem 1 under the control action of (\ref{equ26}) with update laws of (\ref{equ31})-(\ref{equ33}) and (\ref{equ42})-(\ref{equ44}).

\textbf{Proof.} See Appendix B.

\mbox{\textbf{Remark 4.}} It must be clarified that the proposed approach in this study ensures the stability of the interaction in pHRI. Consider the dynamic equations of the exoskeleton in Cartesian space represented at the end-effector frame as,
\begin{figure*}[t]
      \centering
      \subfloat[]{\includegraphics[width = 0.1\linewidth]{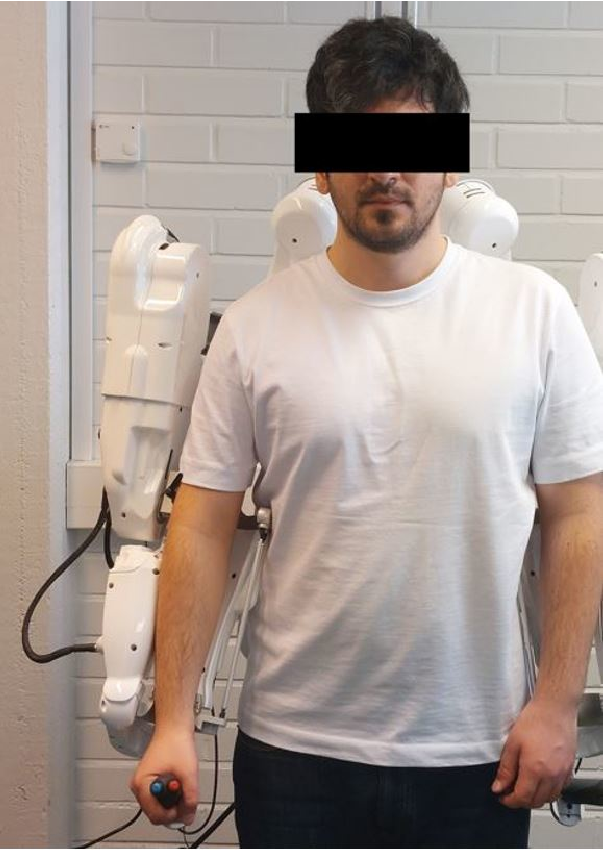}}
      \centering
      \label{fig3a}
      \hfil
      \subfloat[]{\includegraphics[width = 0.1\linewidth]{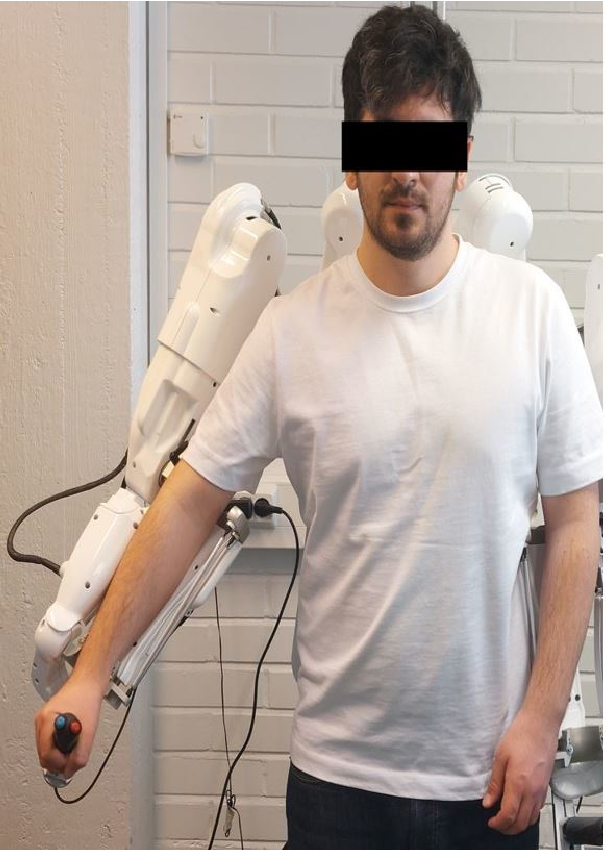}}
      \centering
      \label{fig3b}
      \hfil
      \subfloat[]{\includegraphics[width = 0.1\linewidth]{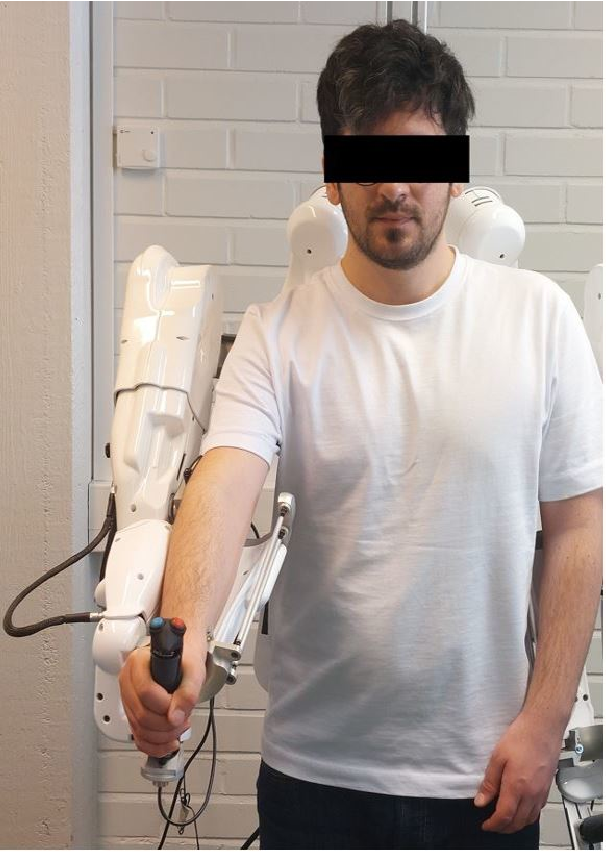}}
      \centering
      \label{fig3c}
      \hfil
      \subfloat[]{\includegraphics[width = 0.1\linewidth]{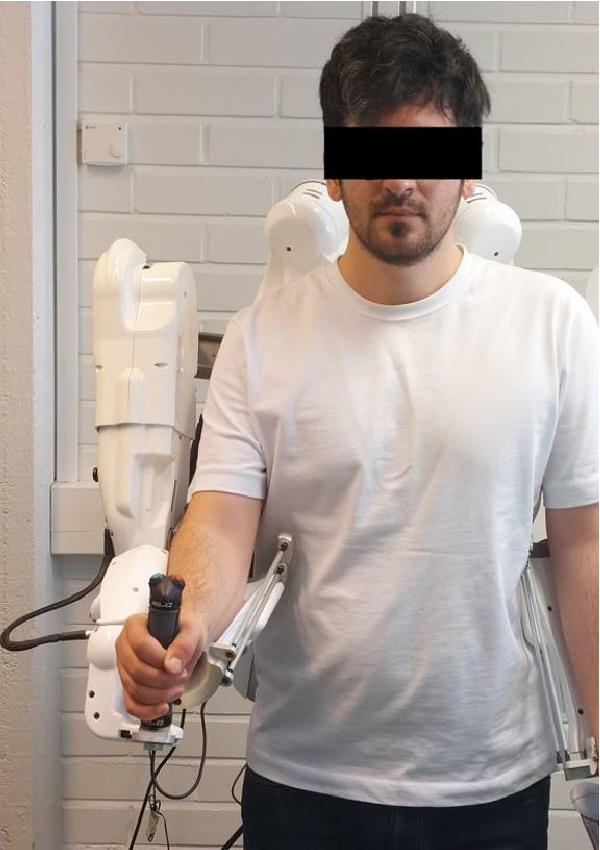}}
      \centering
      \label{fig3d}
      \hfil
      \subfloat[]{\includegraphics[width = 0.1\linewidth]{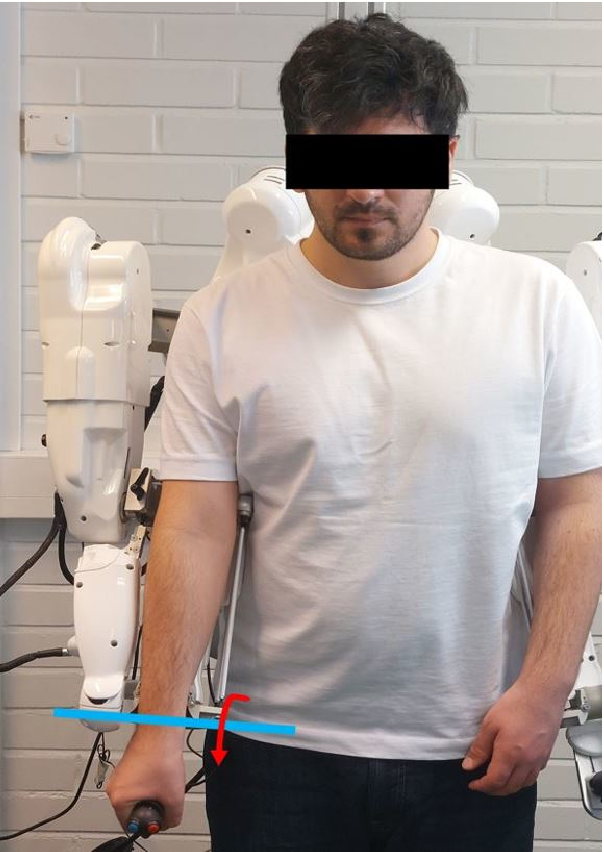}}
      \centering
      \label{fig3e}
      \hfil
      \subfloat[]{\includegraphics[width = 0.1\linewidth]{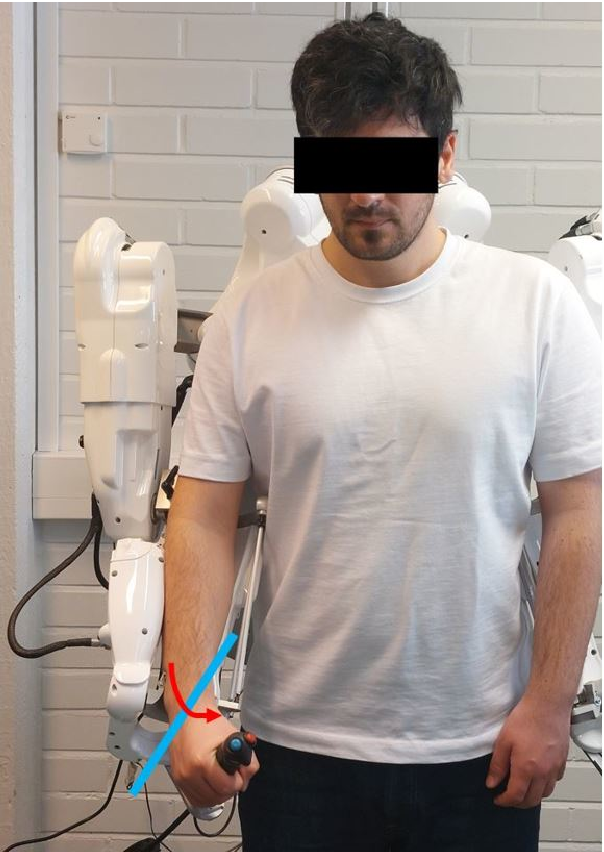}}
      \centering
      \label{fig3f}
      \caption{ Experiment steps for the first part of performance evaluation, a) initial configuration, actuation of b) second, c) third, d) fourth, e) sixth, f) last joint. The joints in actuation work in trajectory tracking mode while others are in set point control mode.}
      \label{Experimental result steps}
   \end{figure*}
   
\begin{figure}[h]
      \centering
      \subfloat[]{\includegraphics[width = 1.65in]{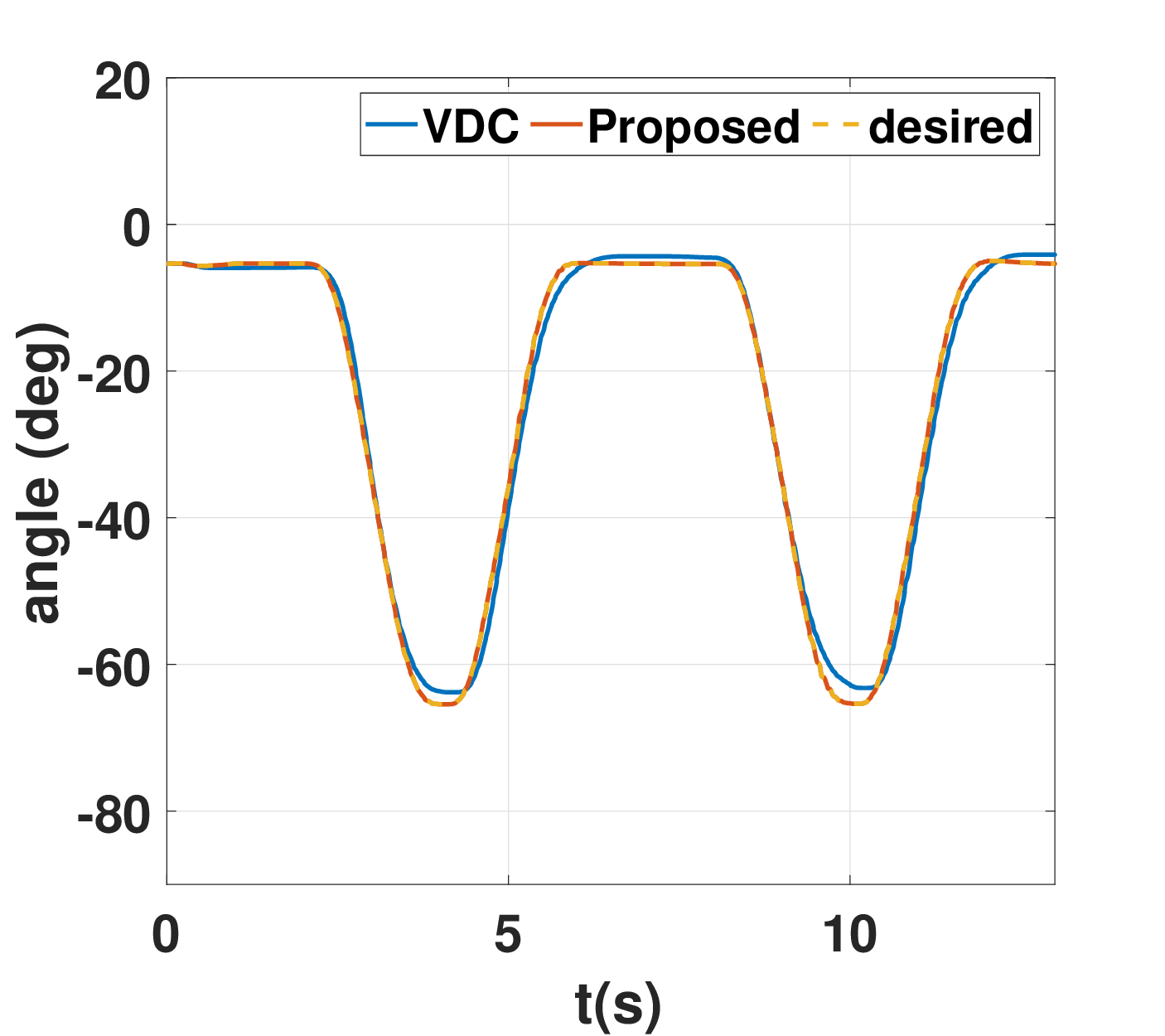}}
      \centering
      \label{fig14}
      \hfil
      \subfloat[]{\includegraphics[width = 1.65in]{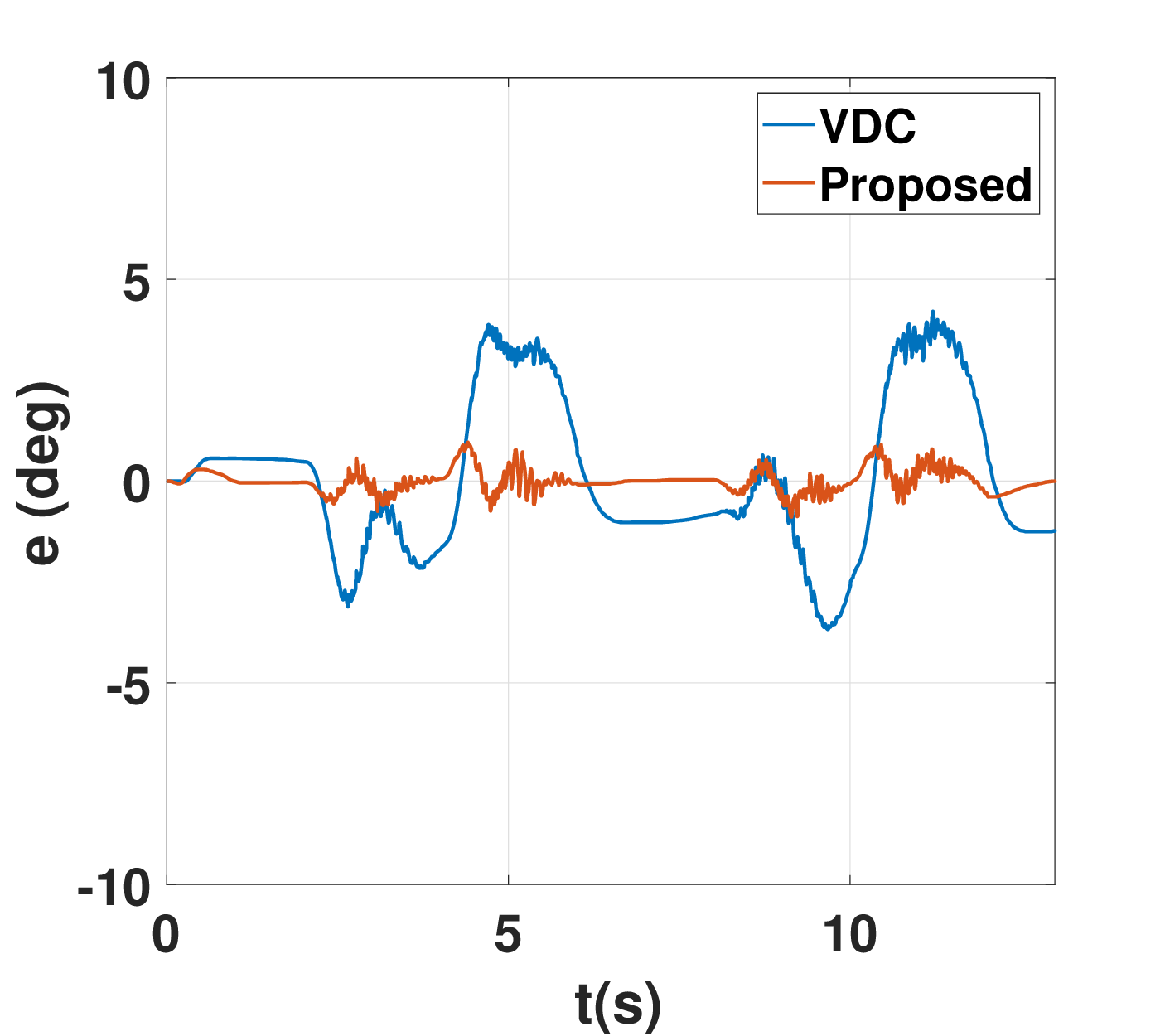}}
      \centering
      \label{fig152}
      \hfil
      \subfloat[]{\includegraphics[width = 1.7in]{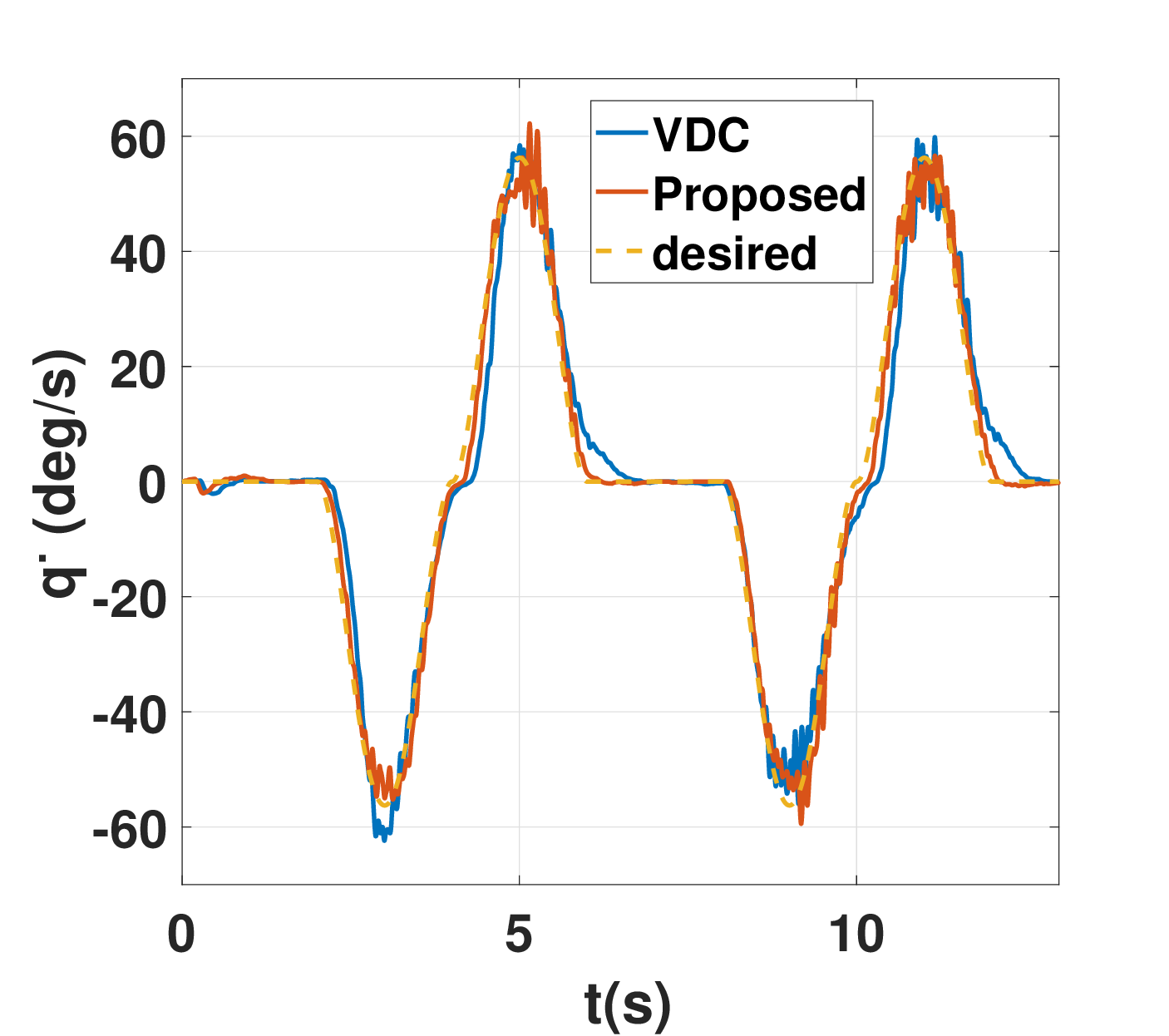}}
      \centering
      \label{fig16}
      \caption{Experimental results of joint number three for comparison of original VDC and proposed controller. a) trajectory tracking, b) tracking error, and c) velocity tracking. The Orange dashed line is the desired trajectory, the solid blue line is for the VDC controller, and the solid red line is for the proposed controller.}
      \label{VDC comparison result}
   \end{figure}
\begin{equation*}
    M_{ex}\Ddot{X}+B_{ex}\Dot{X}+K_{ex}X\, = \, F-F_c
\end{equation*}
where \mbox{\(M_{ex}\)}, \mbox{\(B_{ex}\)}, and \mbox{\(K_{ex}\)} are the Inertia, Damping, and Stiffness matrices of the exoskeleton defined in Cartesian space, and \mbox{\(X\)} is the position of end-effector. \mbox{\(F\)} is the designed force and \mbox{\(F_c\)} is the interaction force between a human and the exoskeleton. Then, the interaction force can be computed as \mbox{\cite{lampinen2021force}},
\begin{equation*}
    F_c = M_h \Ddot{X}_h + B_h \Dot{X}_h + K_h X_h + F_h
\end{equation*}
where \mbox{\(X_h\)} and \mbox{\(F_h\)} are the position and exogenous force of the human arm defined in Cartesian space, respectively, with \mbox{\(M_h\)}, \mbox{\(B_h\)}, \mbox{\(K_h\)} being Inertia, Damping, and Stiffness matrices of the human arm, respectively. Therefore, the interaction force \mbox{\(F_c\)} depends on the inertial parameters of the human arm along with the exogenous force of muscles. In \mbox{(\ref{equ13})} and \mbox{(\ref{equ16})}, both of the mentioned factors have been considered in the control law, presented in \mbox{(\ref{equ26})}. This shows that the presented control law ensures the stability of the interaction in joint space, while passivity-based methods \mbox{\cite{hannaford2002time}} or impedance control methods \mbox{\cite{10210088}} analyses interaction stability at the end-effector frame.

   \begin{figure}[ht]
      \centering
      \includegraphics[width = 2.5in]{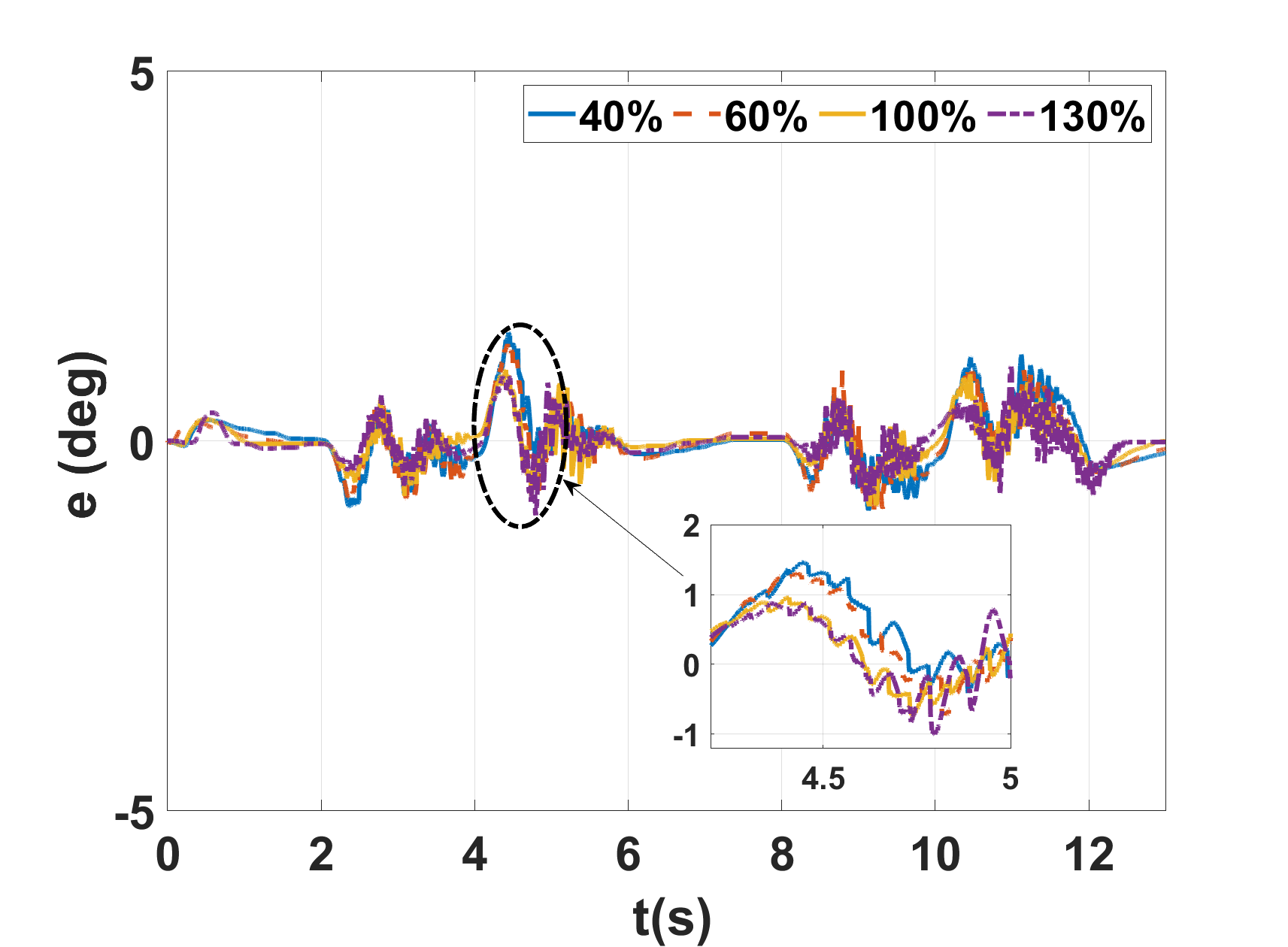}
      \caption{Control parameter sensitivity analysis.}
      \label{paramsensitivity}
   \end{figure}
   
\section{Results}
In this section, the experimental results are provided to evaluate the performance of the proposed neuro-adaptive decentralized controller in comparison to the original VDC, PD, and PID controllers. The Experiments are performed using the commercial 7 DoF upper-limb exoskeleton ABLE \cite{garrec2020design}. All high-level processes are accomplished in a host computer, and computed torque signals are transmitted to the robot using SIMULINK and the Virtuose interface, provided by Haption, with a sample time of 1 ms. The real-time pacer of SIMULINK is utilized to ensure that the experiments will be real-time with the desired sample time. The performance of the presented controller is investigated in four levels. First, the comparison of the presented controller with the original VDC is done. Then, the performance of the controllers is evaluated in three different trajectories and velocities (Fig. 4), with three different operators (Fig. 10), and with unknown disturbances (Fig. 11). A 5th-order acceleration-smooth trajectory generator \cite{jazar2010theory} is utilized to produce a desired trajectory between two points. This generator gets the initial and final point and generates a smooth trajectory based on task execution time \(t_f\), indicating the velocity. The smaller \(t_f\), the faster the trajectory. The radial basis function for RBFNN in (\ref{equ28}) and (\ref{equ40}) is chosen as a Gaussian function described as \(\Psi_{\mathbf{j}}(\chi) = exp([-(\chi-c_{\mathbf{j}})^T(\chi-c_{\mathbf{j}})/(b_{\mathbf{j}}^2)])\), with \(c_{\mathbf{j}}\) and \(b_{\mathbf{j}}\) denoting the center and width of the neural cell in \(\mathbf{j}\)th unit. In this study, the value for \(c_{\mathbf{j}}\) is randomly selected in \([-1,1]\) and \(b_{\mathbf{j}} = 1\) for all the 500 nodes.

In order to better evaluate and compare the proposed controller with the original VDC, the experimental result of controlling a single joint of the exoskeleton (joint number three) is reported. Control gains are set to: \mbox{\(\lambda_3 = 5,\, \mathcal{K}_{D3} = 2,\, \mathcal{K}_{I3} = 4,\, k_{d3} = 0.5,\, k_{I3} = 4,\, k_{b3} = 5\)}. Fig. 5 demonstrates the tracking performance of both controllers. As can be seen from Fig. 5(a) and 5(b), the tracking error with the proposed controller is considerably lower than the original VDC given the same control gains. Fig. 5(c), also, displays the desired joint velocity tracking. It can be concluded that even for such a fast motion (around 60 deg/s), the proposed controller perfectly follows the desired trajectory. The root mean square (RMS) value of the tracking error for the original VDC and proposed controller is 1.92 deg and 0.28 deg, respectively. Such an excellent performance happens when both controllers require approximately the same torque (the RMS torque value for the original VDC and presented controller is 2.04 and 2.14, respectively). Additionally, in order to evaluate the effect of control gains over the stability and performance of the proposed controller, the tracking error with \mbox{\(40\%\)}, \mbox{\(60\%\)}, \mbox{\(100\%\)}, and \mbox{\(130\%\)} of the above-mentioned control gains are provided in Fig. 6. The RMS value of \mbox{\(40\%\)}, \mbox{\(60\%\)}, \mbox{\(100\%\)}, and \mbox{\(130\%\)} is 0.39, 0.37, 0.28, and 0.27. All these show that the proposed controller is robust against control parameters. In the following, the result of the presented controller is compared to the PD and PID controllers.

\begin{figure*}[t]
      \centering
      \subfloat[]{\includegraphics[scale = 0.22]{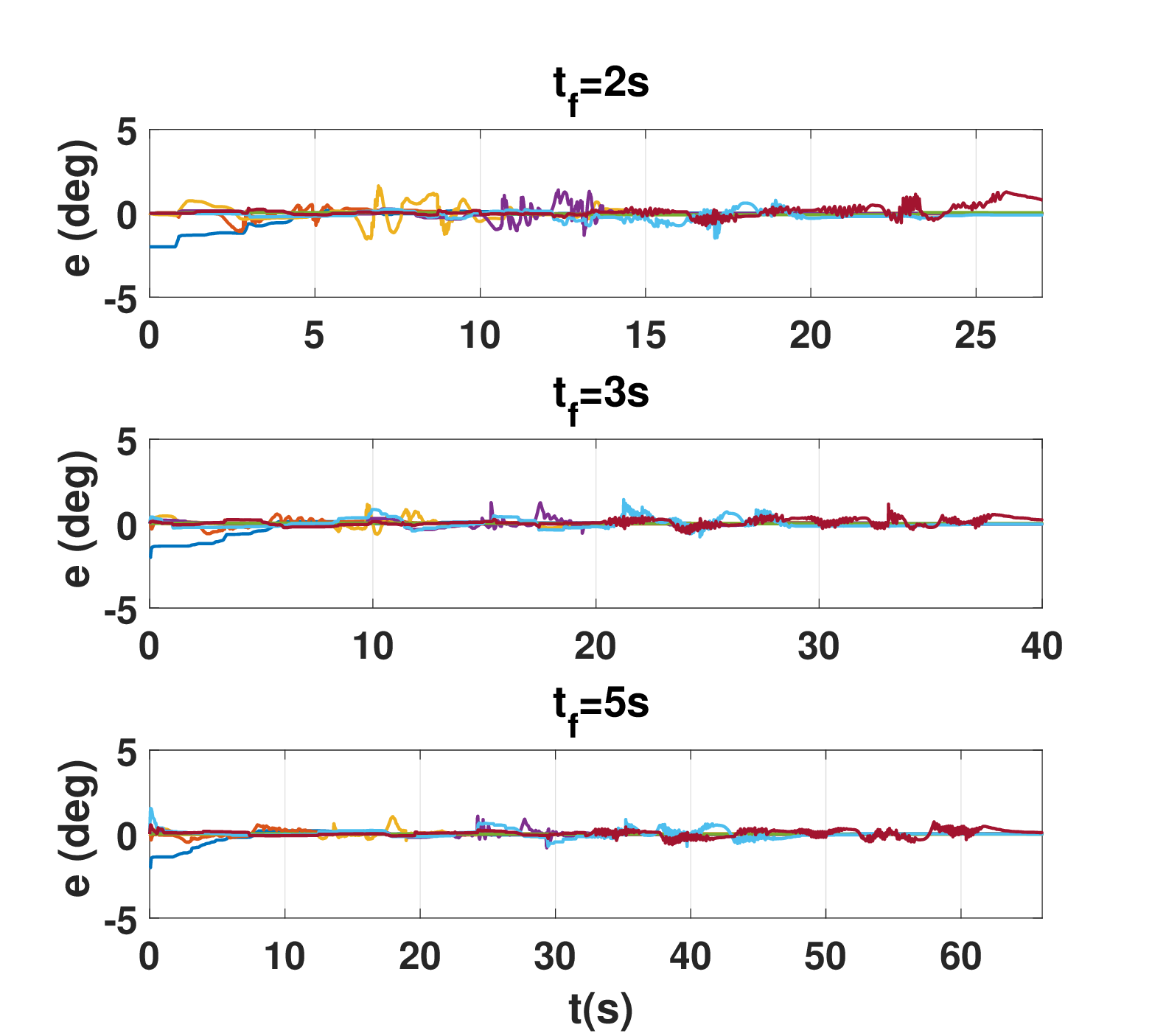}}
      \centering
      \label{fig44}
      \hfil
      \subfloat[]{\includegraphics[scale = 0.22]{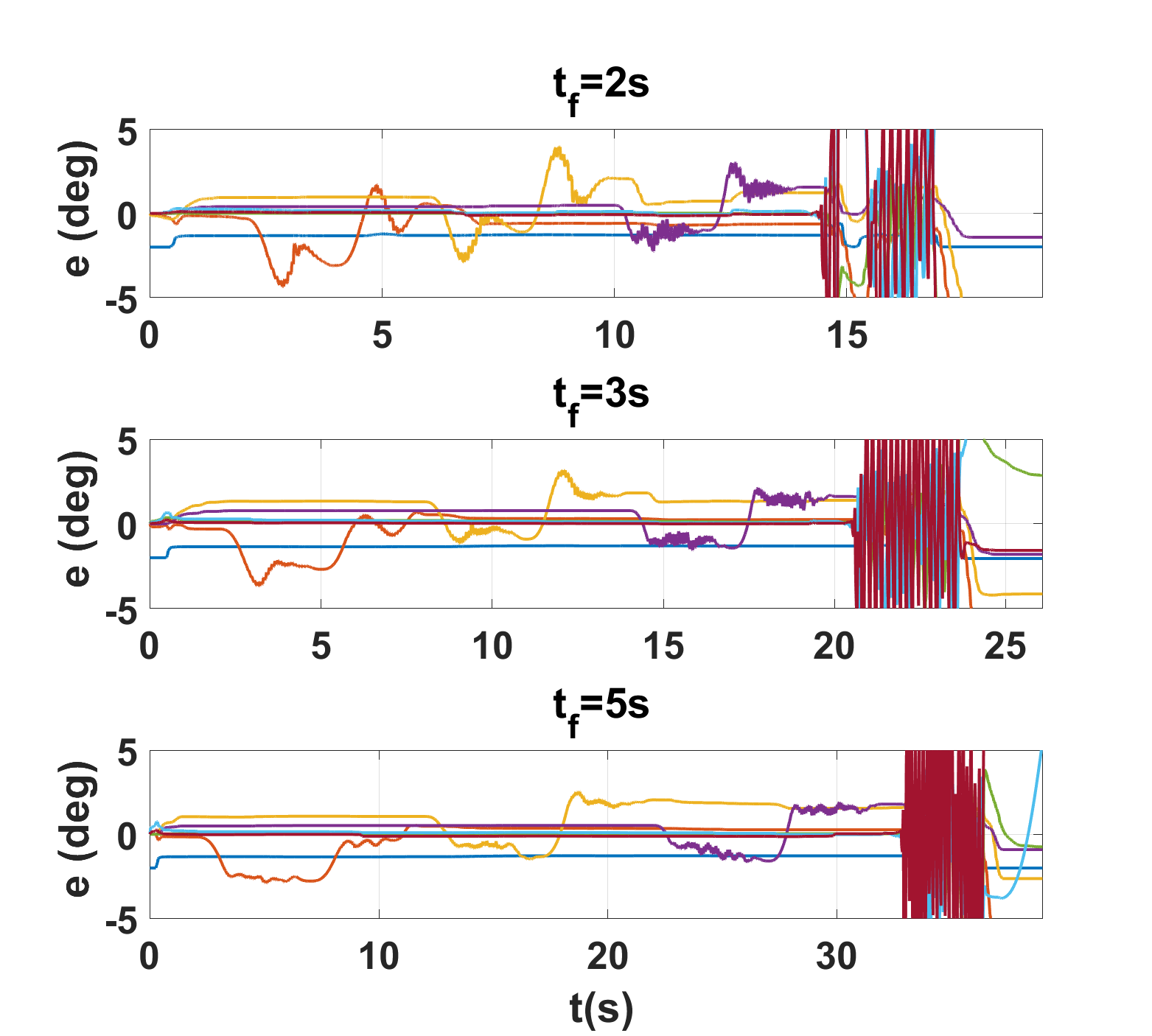}}
      \centering
      \label{fig54}
      \hfil
      \subfloat[]{\includegraphics[scale = 0.22]{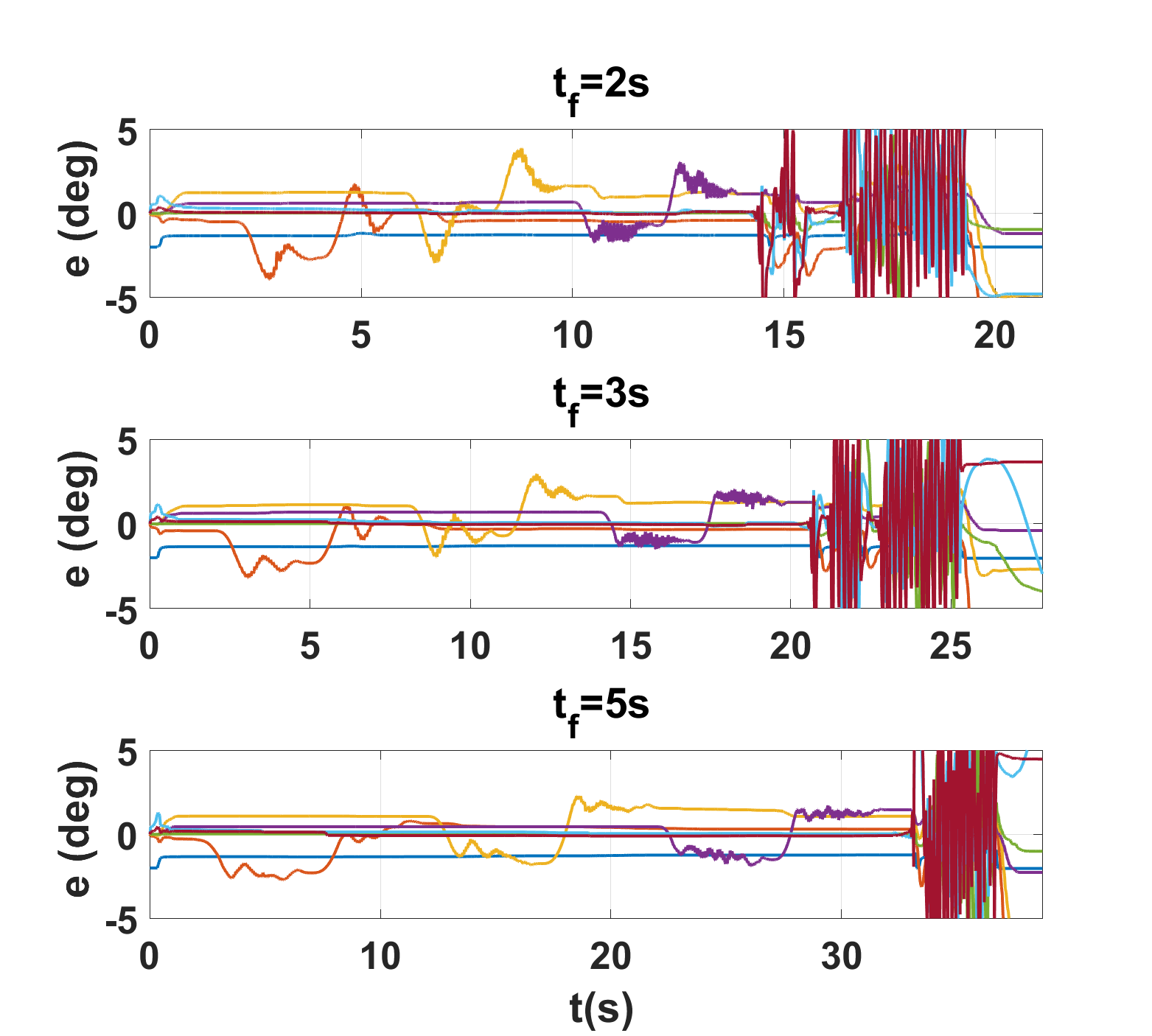}}
      \centering
      \label{fig6}
      \caption{Tracking error of each joint in experiments with different controllers and trajectories. a) proposed controller, b) PD controller, c) PID controller. Each color in the figures represents the joint number.}
      \label{Experimental result2}
   \end{figure*}
   
The PD and PID controller are defined as,
\begin{equation}\label{equ49}
    \tau_{PD}(t) = k_{pi}(q_{di}-q_i(t))+k_{vi}(\Dot{q}_{di}-\Dot{q}_i(t))
\end{equation}
\begin{equation}\label{equ50}
    \tau_{PID}(t) = \tau_{PD}(t) + \Bar{k}_{Ii}\, \int_{0}^{t} (\Dot{q}_{di}-\,\Dot{q}_i) dt
\end{equation}
where \(k_{pi}\), \(k_{vi}\), and \(\Bar{k}_{Ii}\) are positive constants. To evaluate the performance of each controller, the performance index is defined as \cite{zhu2005adaptive},
\begin{equation}\label{equ51}
    J = \max(|e|)/\max(|\Dot{q}|)
\end{equation}
where \(|.|\) is the absolute value function and joint tracking error defined as \(e(t) = q_d(t)-q(t)\). In contrast to other indices that are based on error, the chosen index computes the ratio of the maximum error and maximum velocity. If a controller achieves a very small error in a very slow motion, the index will show a bigger number than when the controller achieves a small error in high velocity. Undoubtedly, having a small error in faster motion is desired and demonstrates the perfect performance of the controller. The control parameters utilized for this part of the experiments are presented in Table. 1. The constraint parameters in (\ref{equ41}) are as, \(k_c = [10,\, 30,\, 40,\, 60,\, 25,\, 25,\, 25]^T (deg)\), \(k_{cr} = [15,\, 35,\, 48,\, 68,\, 40,\, 40,\, 40]^T (deg)\), which result in error constraint as \(k_b = [5,\, 5,\, 8,\, 8,\, 15,\, 15,\, 15]^T (deg)\). The saturation level for the first four motors is \(\pm5N\cdot m\) and for the last three joints \(\pm1.2N\cdot m\). The deadzone parameters are considered unknown. 

\begin{table}[ht]
\centering
\caption{Control parameters for experiment}
\label{table}
\setlength{\tabcolsep}{10pt}
\begin{tabular}{|p{25pt}|p{25pt}|p{25pt}|p{25pt}|}
\hline
Symbol& 
Value& 
Symbol& 
Value
\\
\hline
$\lambda_i $& 
0.5& 
$\mathcal{K}_{D1,2} $&
0.1I$_{6\times6}$\\
$\mathcal{K}_{D3,4} $&
0.5I$_{6\times6}$&
$\mathcal{K}_{D5} $&
03I$_{6\times6}$\\
$\mathcal{K}_{D6,7} $&
0.2I$_{6\times6}$&
$\mathcal{K}_{I1} $&
5I$_{6\times6}$\\
$\mathcal{K}_{D2} $&
3I$_{6\times6}$&
$\mathcal{K}_{D3,4} $&
2I$_{6\times6}$\\
$\mathcal{K}_{I5,6} $&
2I$_{6\times6}$ &
$\mathcal{K}_{I7} $&
0.2I$_{6\times6}$ \\
$\Gamma_{Bi} $& 
2I$_{9\times9}$&
$\gamma_1 $&
0.1 \\
$\gamma_{2i} $& 
1 &
k$_{d1,2}$ &
0.05 \\
k$_{d3,4}$ &
0.2 &
k$_{d5,6,7}$ &
0.01 \\
k$_{I1}$ & 
5 &
k$_{I2}$ & 
3 \\
k$_{I3,4}$ & 
2 &
k$_{I5}$ & 
0.2 \\
k$_{I6,7}$ & 
0.1 &
$\zeta $&
1 \\
$\beta_{1i} $& 
0.5 &
$\beta_{2i} $&
1 \\
k$_{pi}$ & 
50 &
k$_{vi}$ & 
0.5 \\
$\Bar{k}_{Ii}$ & 
1 &
 & 
 \\ 

\hline
\end{tabular}
\label{tab}
\end{table}

\begin{figure}[ht]
      \centering
      \subfloat[]{\includegraphics[width = 1.7in]{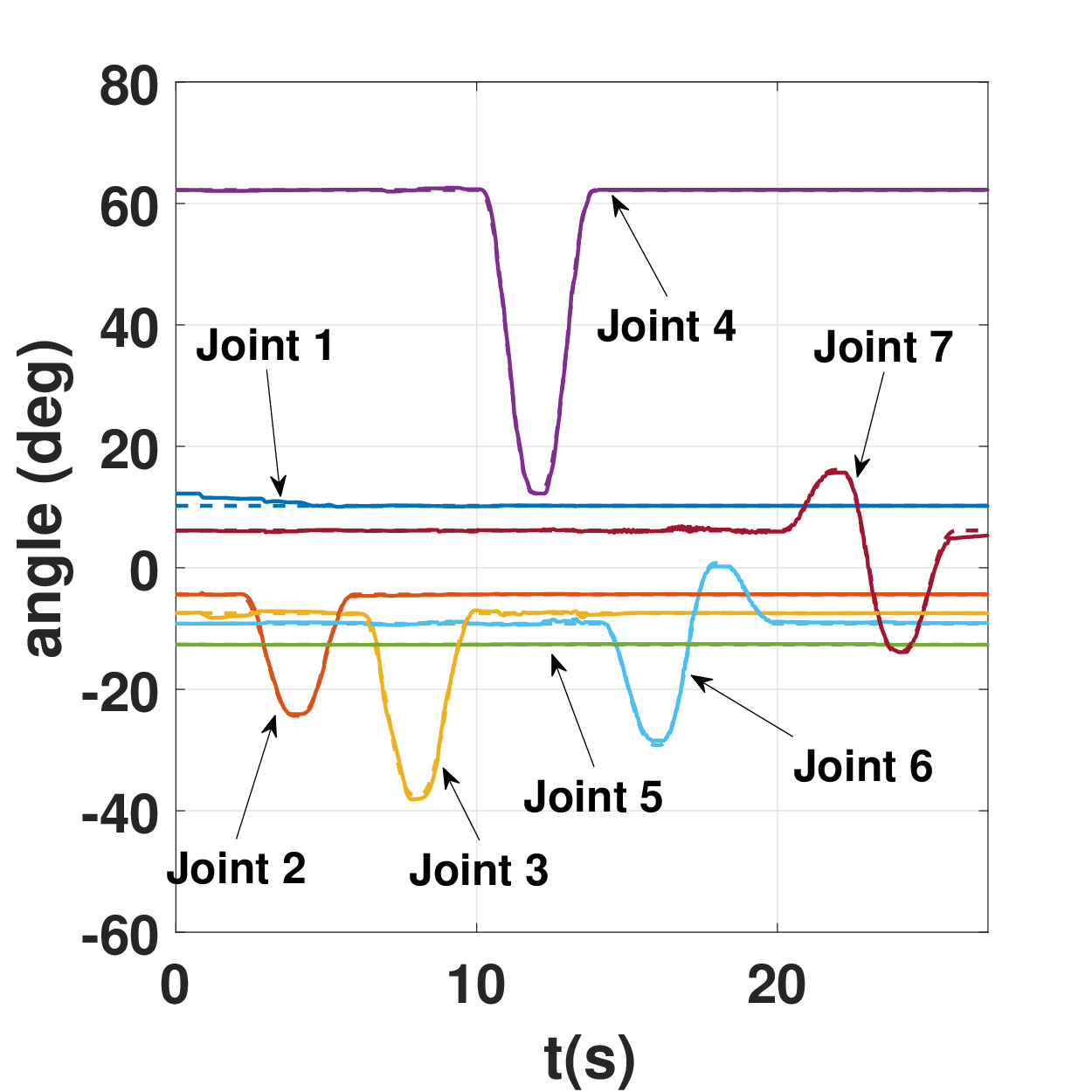}}
      \centering
      \label{fig124}
      \hfil
      \subfloat[]{\includegraphics[width = 1.7in]{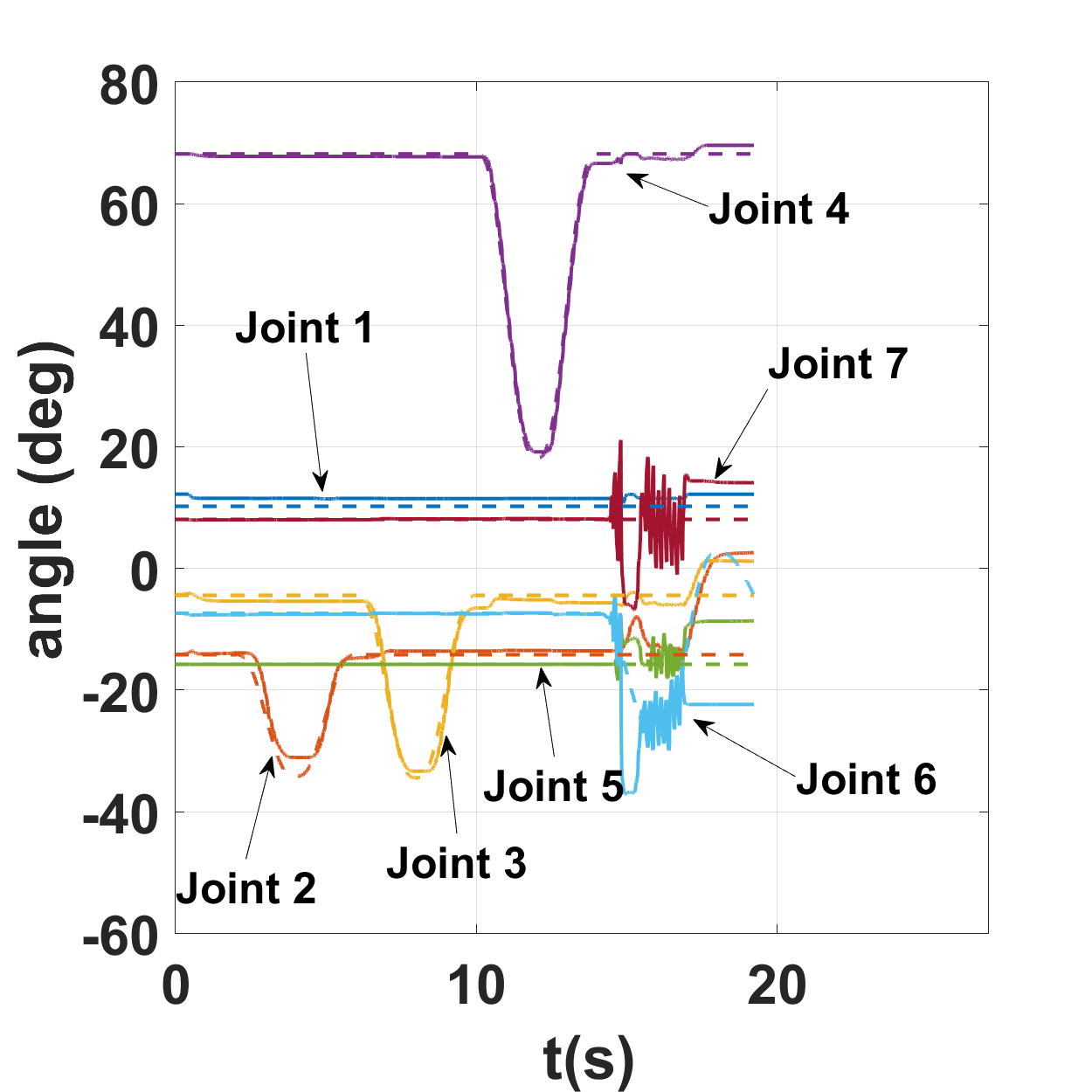}}
      \centering
      \label{fig15}
      \hfil
      \subfloat[]{\includegraphics[width = 1.7in]{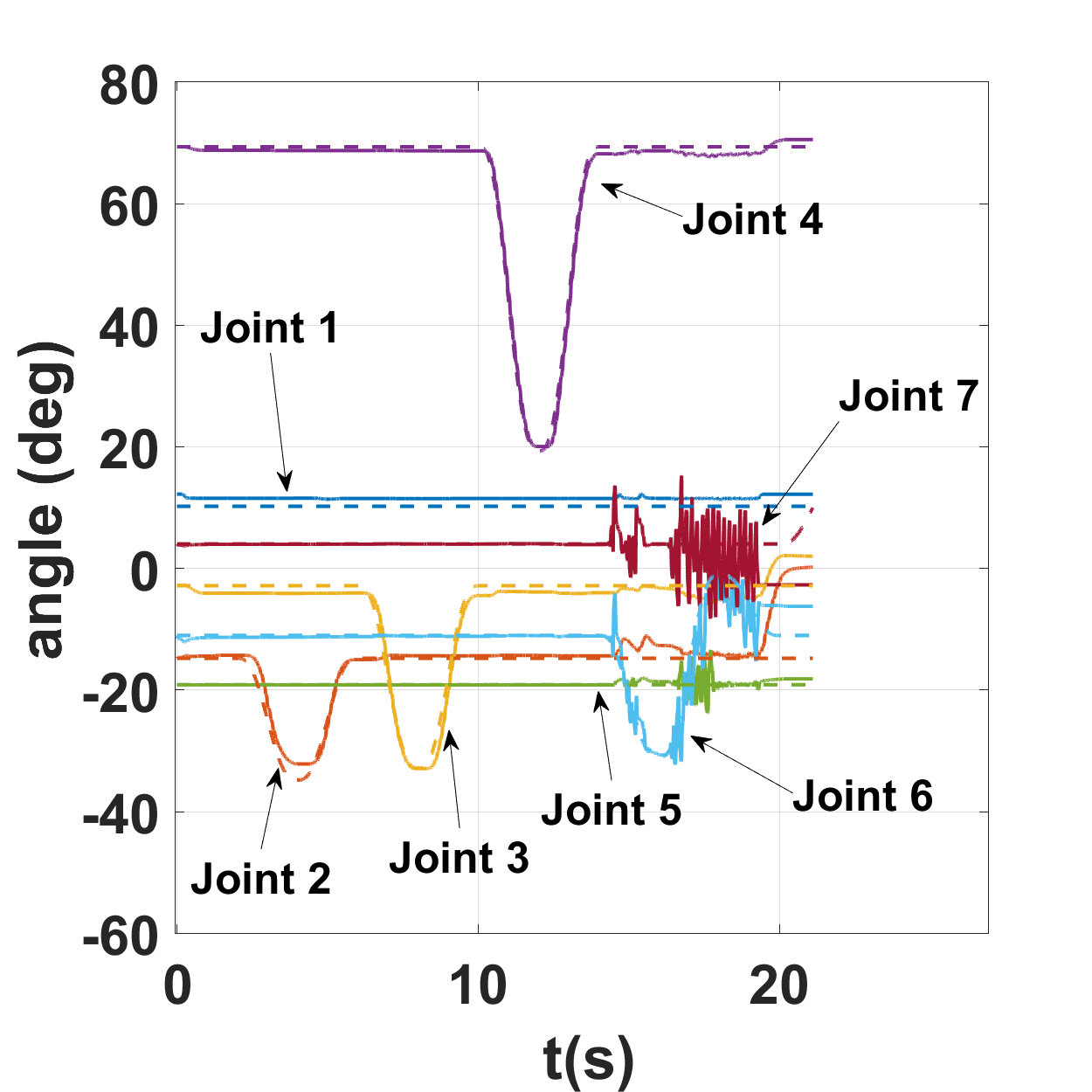}}
      \centering
      \label{fig156}
      \caption{Experimental results of trajectory tracking of each joint with \(t_f = 2s\). a) Proposed controller b) PD controller, and c) PID controller. The dashed line is the desired trajectory and the solid line is the actual signal. Due to instability in joints 6 and 7, experiments are terminated sooner in (b) and (c).}
      \label{Exp Control result proposed}
   \end{figure}

   \begin{figure}[ht]
      \centering
      \subfloat[]{\includegraphics[width = 1.7in]{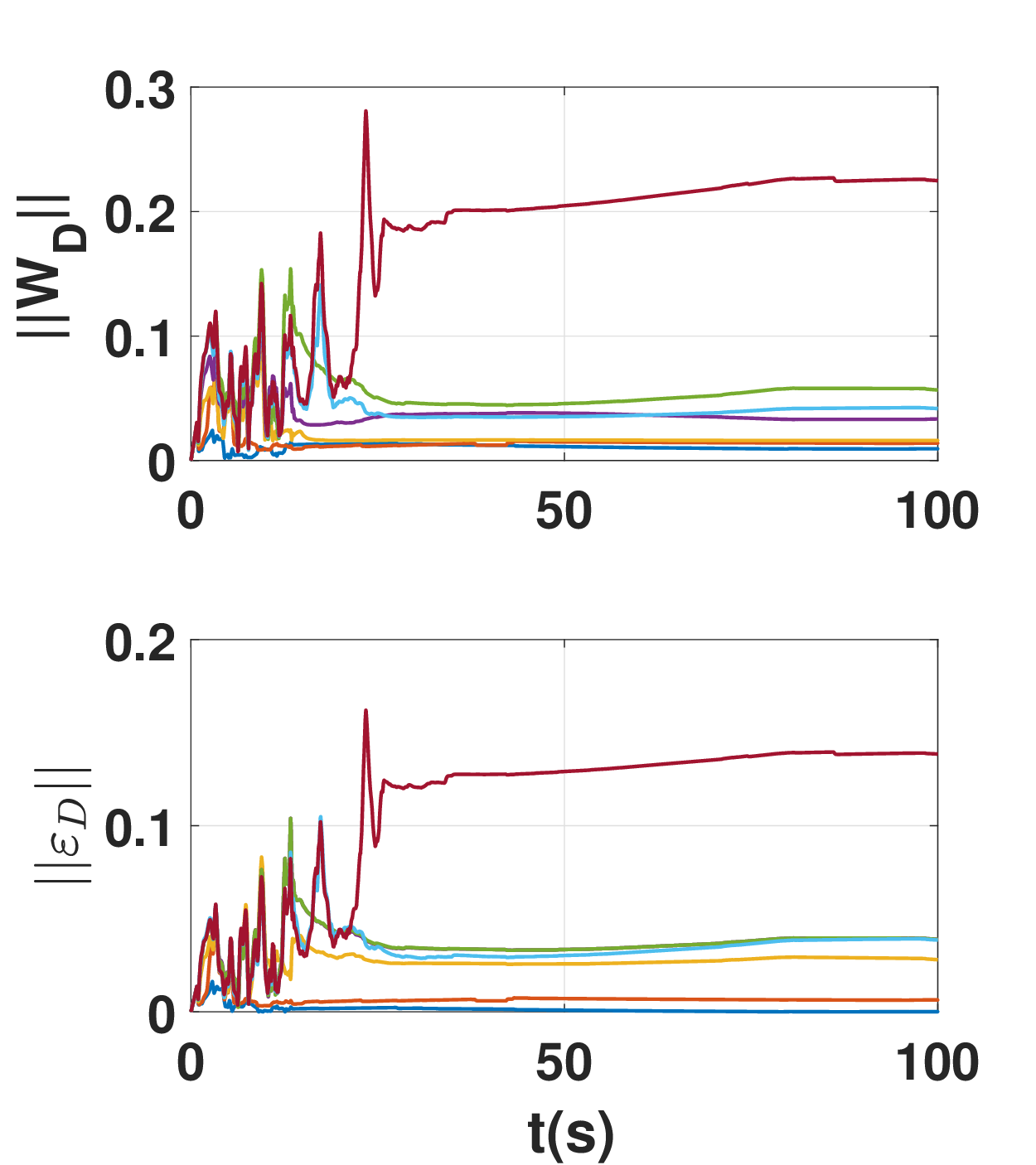}}
      \centering
      \label{fig64}
      \hfil
      \subfloat[]{\includegraphics[width = 1.7in]{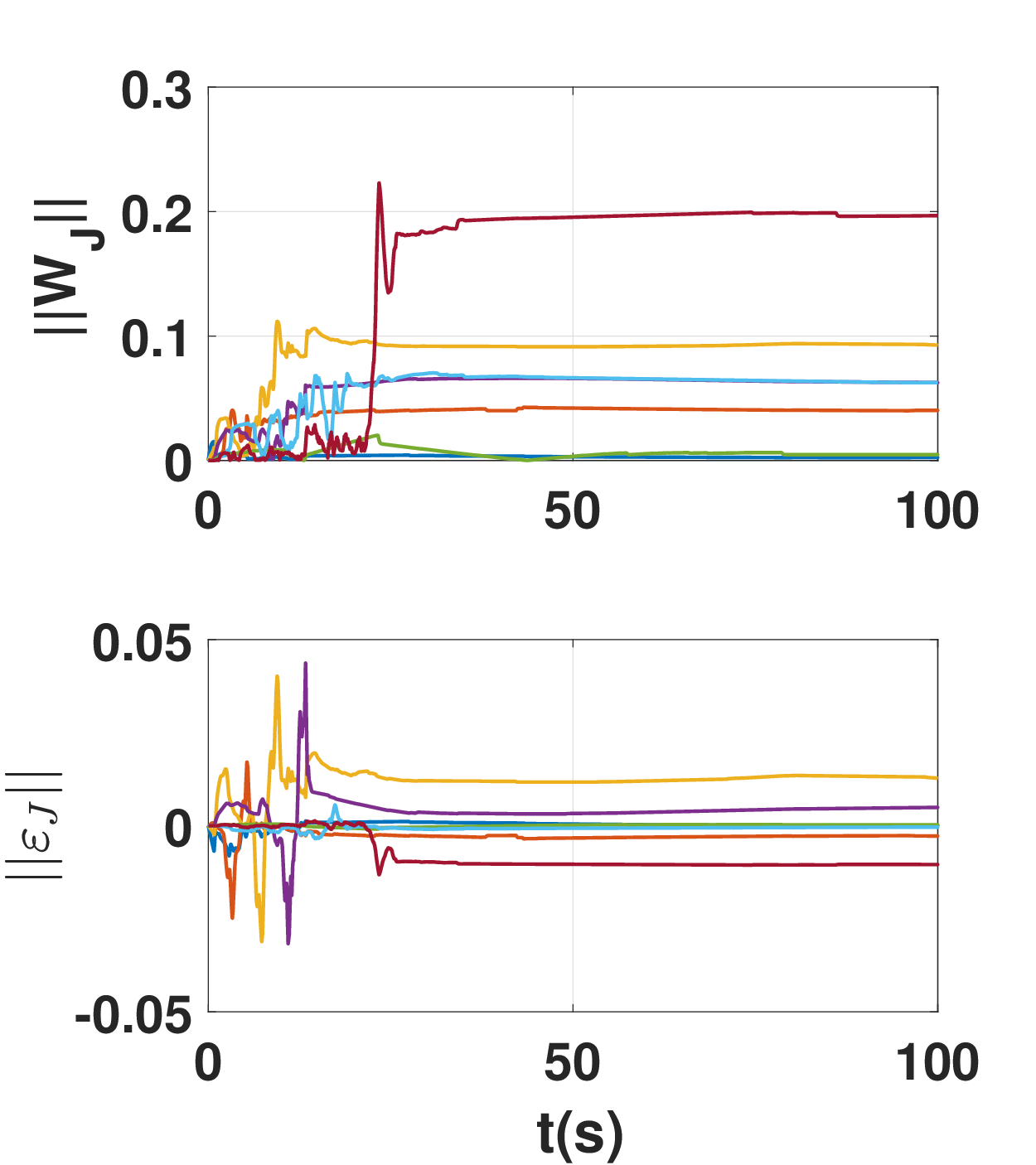}}
      \centering
      \label{fig65}
      \hfil
      \subfloat[]{\includegraphics[width = 1.7in]{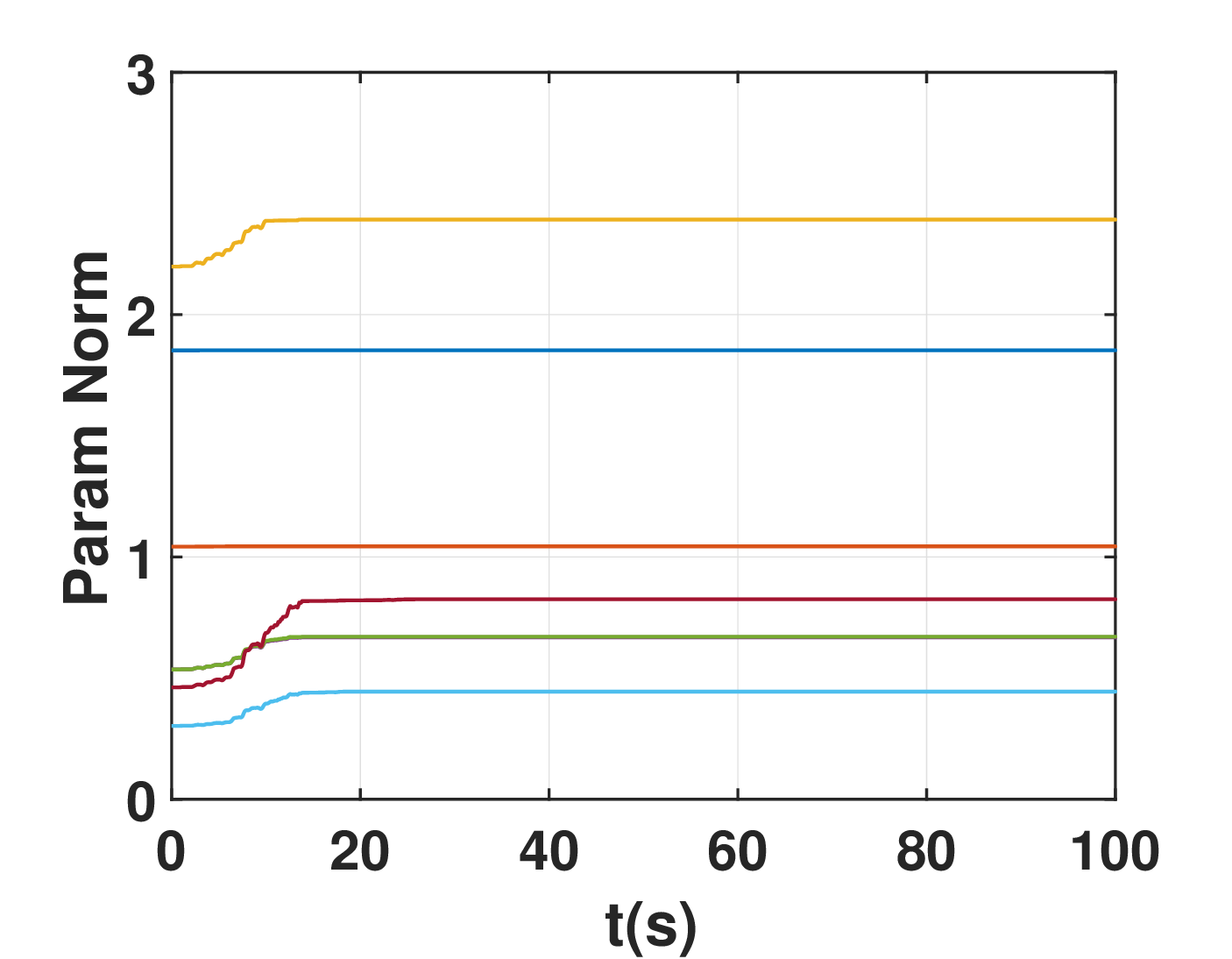}}
      \centering
      \label{fig66}
      \caption{Time history of parameter adaptation. RBFNN parameter estimation is shown in a) rigid body, and b) joint part. c) Norm of inertial parameter estimation of the entire exoskeleton.}
      \label{Exp Control result proposed6}
   \end{figure}

Fig. 7 depicts the experimental results of tracking error for all three controllers with different trajectories. Fig. 7(a), 7(b), and 7(c) show the trajectory tracking error under the proposed, PD, and PID controllers in three different velocities, respectively. As shown in Fig. 7(b) and 7(c), PD and PID controllers could not show a good performance, especially in control of the last two joints (wrist) that are driven by ball screws and cable. As a result, steps e and f in Fig. 4 are terminated without accomplishment due to the instability of the controllers. In contrast, not only the proposed controller could stabilize the last two joints but also reached a tracking error of less than two degrees. This is because the unknown dynamics of the rigid body and actuator part are learned through RBF networks, enabling the proposed controller to handle uncertainties.  Moreover, as it can be seen from Fig. 7(a), the tracking errors of all joints with different angular velocities are less than the error limit, \(k_b\), ensuring safety despite joint angular velocity. Fig 8, in addition, demonstrates the trajectory tracking with \(t_f = 2s\) for three controllers. Instability in the last two joints is shown in Fig. 8(b) and 8(c) which resulted in the termination of the experiment. However, the perfect performance of the presented controller can be seen in Fig. 8(a). The RMS value for the performance index \(J\), defined in (\ref{equ51}), of all joints related to Fig. 8 is 0.625, 0.563, and 0.0068 for PD, PID, and proposed controller, respectively. To better evaluate the performance of the controllers, RMS of joint tracking errors (RMSE) needs to be provided as well. The maximum RMSE of all seven joints are 2.95, 3.08, and 0.26 for PD, PID, and VDC, respectively. The provided metrics perfectly show that the proposed controller has demonstrated good performance in accomplishing the objectives.

Fig. 9 depicts the norm of updated neural network parameters along with the estimation of unknown rigid body parameters. Fig. 9(a) shows the norm of the updated weights and error vector for all the rigid bodies. Fig. 9(b) displays the norm of the updated weights and error vector for all the joints. Fig. 9(c) demonstrates the norm of the estimated unknown inertial parameters of all rigid bodies. As these figures illustrate, all the norms of estimated parameters are bounded during the experiments. Therefore, the stability of the robot and all states under the designed controller is validated experimentally.

\begin{figure}[ht]
      \centering
      \includegraphics[width = 3in]{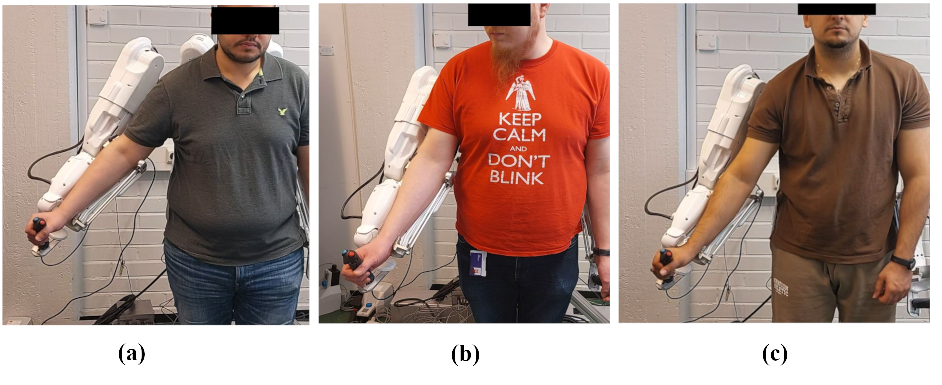}
      \caption{Experiments of different human operators. a) subject 1, b) subject2, and c) subject 3.}
      \label{figurelabel7}
   \end{figure}

   \begin{table}[ht]
\caption{Performance evaluation of controllers with different human users with \(t_f = 2s\)}
  \begin{tabular}{|c|c|c|c|c|c|c|}
  \hline
    \multirow{2}{*}{Controller} &
      \multicolumn{2}{c|}{Subject 1} &
      \multicolumn{2}{c|}{Subject 2} &
      \multicolumn{2}{c|}{Subject 3} \\
      \cline{2-7}
      & {\(J_{rms}\)} & {\(\tau_{rms}\)} & {\(J_{rms}\)} & {\(\tau_{rms}\)} & {\(J_{rms}\)} & {\(\tau_{rms}\)} \\
      \hline
    PID & 0.66 & 1.83 & 0.549 & 2.49 & 0.633 & 1.52 \\
    VDC & 0.0108 & 2.19 & 0.009 & 3.112 & 0.0105 & 3.042 \\
    \hline
    
  \end{tabular}
\end{table}

Table 2 demonstrates the RMS value for \(J\) and torque for PID and the proposed controller subjected to different human operators (Fig. 10). The detail of each user is as follows. Subject 1: 28 years old, 175 cm, 110 kg. Subject 2: 30 years old, 190 cm, 120 kg. Subject 3: 31 years old, 186.5 cm, 90 kg. None of the subjects are informed about the robot's motion which allows us to evaluate the performance of controllers in the existence of human arm resistance. As can be seen from Table 2, the performance of the designed controller is much better than the PID controller. Such a perfect performance is due to the incorporation of HET and human arm parameter estimation into the VDC control scheme. To achieve the results in Table 2 for performance evaluation, only the second and third joints are in trajectory tracking mode and others are only commanded to stay in the initial configuration. This is because the PID controller can not stabilize the wrist joints (last two joints) and it is not safe for users. 

Fig. 11 demonstrates the set-up for examining the performance of each controller in the presence of unknown disturbance. Disturbances are applied by a human to different locations on the robot arm with unknown directions and magnitudes. The results are displayed in Fig. 12 for PID and the proposed controller. As can be seen from Fig. 12(a), PID control could not tolerate unknown disturbances and got unstable with huge errors. In contrast, the presented controller (Fig. 12(b)) ensured the stability of the robot in the presence of such an unknown disturbance with unknown magnitude and direction. It can be seen from Fig. 12(b) that the proposed controller guaranteed the state constraints despite unknown disturbances, which shows the excellent result of incorporating RBFNNs and BLF into the VDC scheme. The mentioned results perfectly demonstrate the robustness of the proposed controller in different scenarios, ensuring the safety of the human operators and robot during pHRI. It is worth mentioning that the run-to-sim ratio provided by SIMULINK for the performed experiments is 1.08. This perfectly shows that despite RBFNNs, the experiments are performed in real-time.

\begin{figure}[ht]
      \centering
      \includegraphics[scale = 0.25]{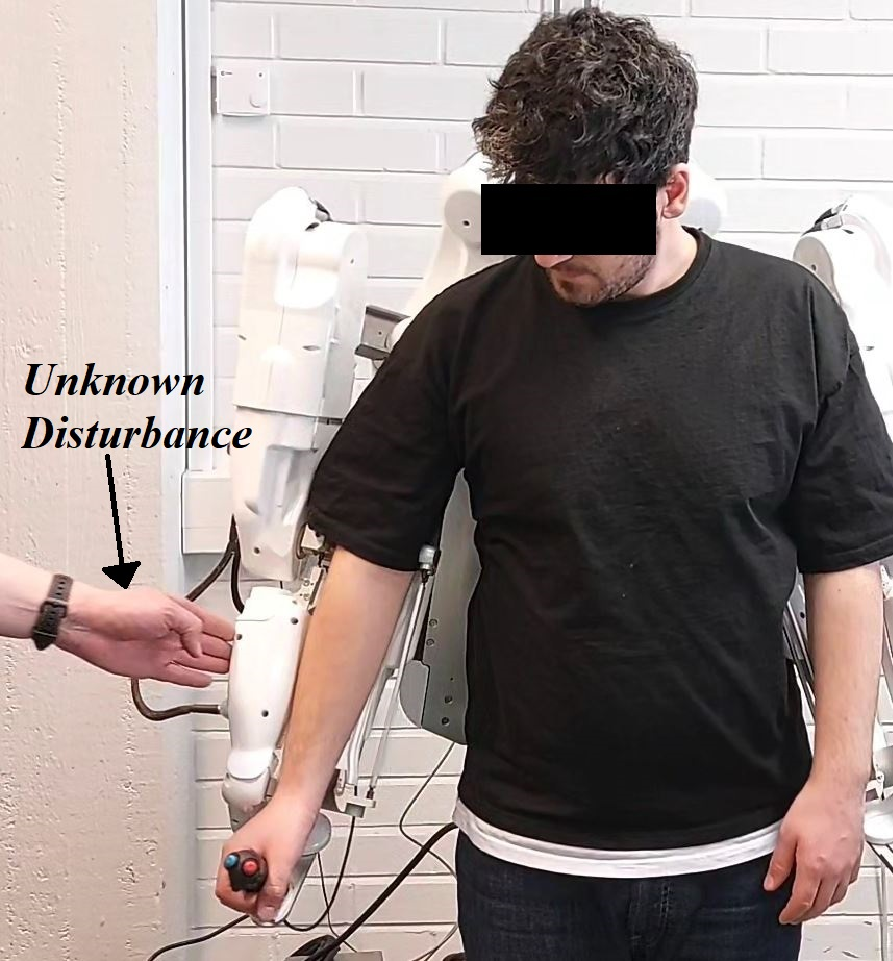}
      \caption{Experimental set-up for disturbance rejection capability analysis}
      \label{figurelabel81}
   \end{figure}

      \begin{figure}[ht]
      \centering
      \subfloat[]{\includegraphics[width = 2.5in]{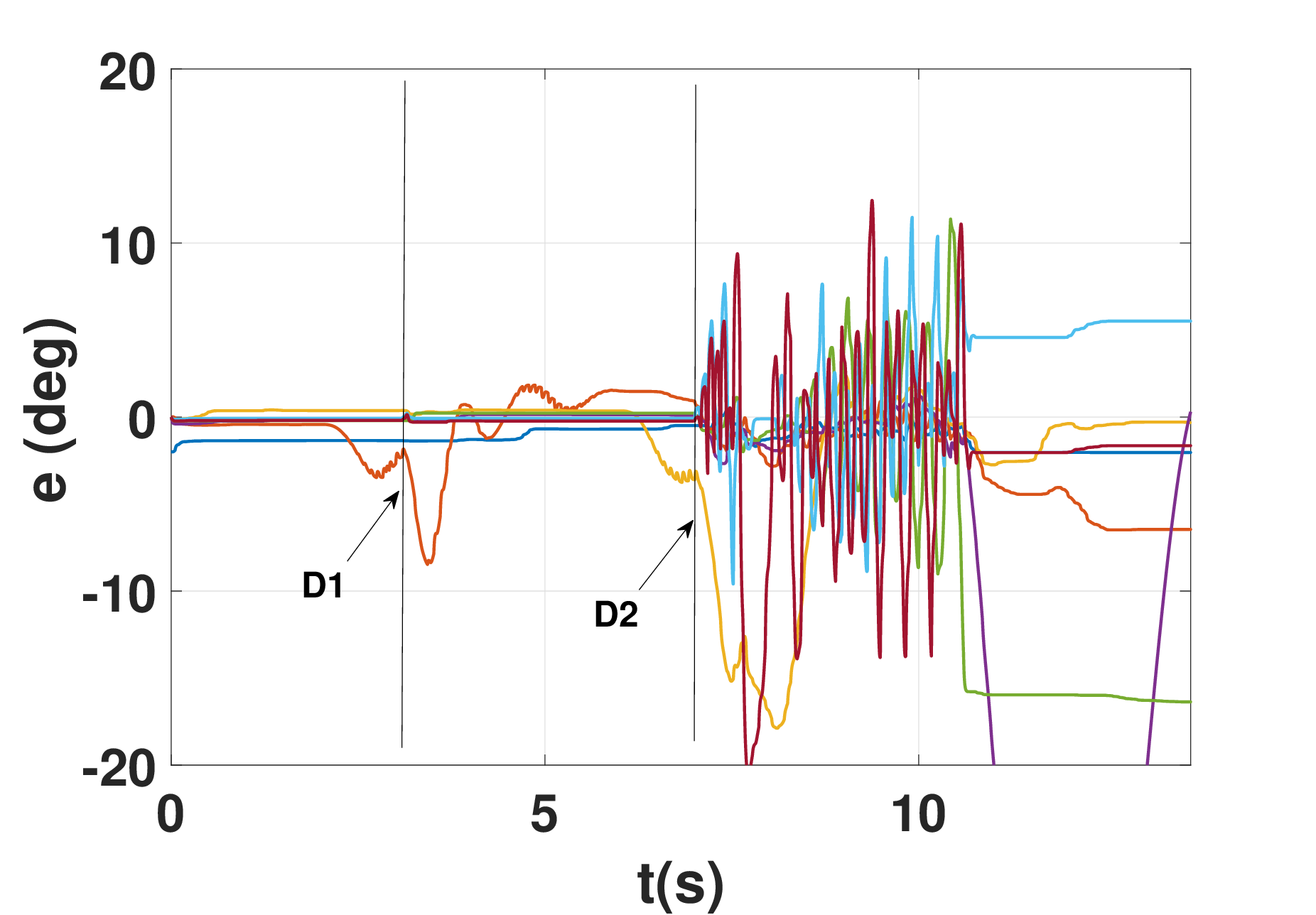}}
      \centering
      \label{fig91}
      \hfil
      \subfloat[]{\includegraphics[width = 2.5in]{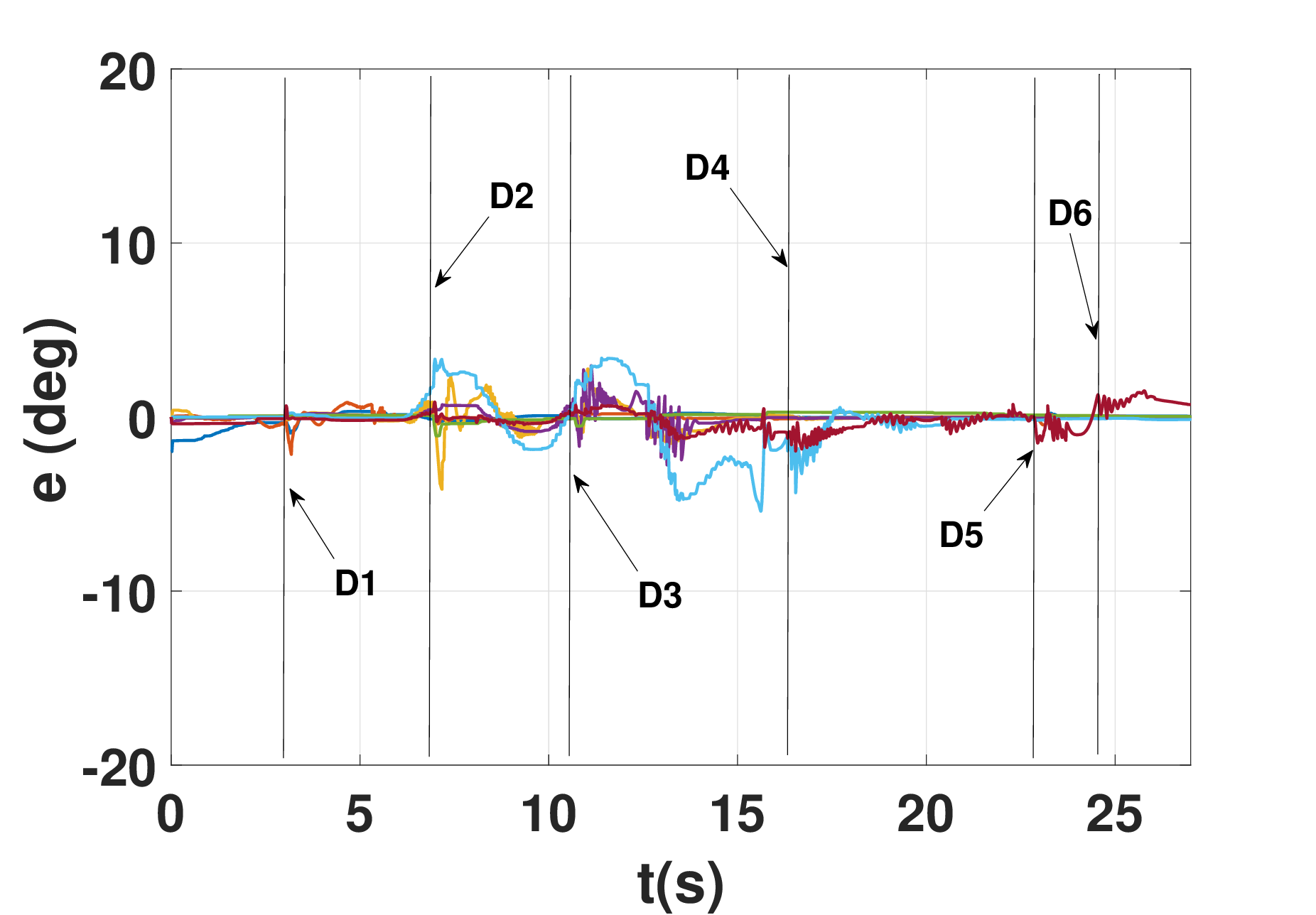}}
      \centering
      \label{fig92}
      \caption{ Tracking error of all joints of the robot under unknown disturbances. a) with PID controller, b) with proposed controller. Each color demonstrates a joint.}
      \label{Exp Control result proposed9}
   \end{figure}

\section{Conclusion}
Considering the safety of human operators in the physical human-robot interaction of upper limb exoskeleton, this study designed a robust RBFNN-based adaptive decentralized controller that ensures the safety of human operators by considering unknown robot and human model uncertainty, actuator dynamics constraints and uncertainties. The employed VDC scheme helped to break down the entire complex system into subsystems and design the controller at the subsystem level. Estimating HET and inertial parameters of the human arm and considering them in the control law increased the performance of the designed controller and made it robust and adaptable to different human operators. In order to apply constraints on states and guarantee physical safety, the control law is modified using the barrier Lyapunov function. It is shown that having BLF for state constraint along with RBFNNs for estimating unknown disturbance and model uncertainty increased the robustness and performance of the proposed controller in comparison to the original VDC. The stability of the entire system under the designed control law is analyzed by means of virtual stability and VPFs, and asymptotic stability is established. Finally, experimental results are performed to examine the robustness and performance of the proposed controller. As is demonstrated in the results section, the presented controller demonstrated excellent performance in stabilizing the system and accomplishing the control objectives in the presence of unknown uncertainties and disturbances in comparison to PD and PID controllers.

\section{appendix}

\subsection{Defining a unique symmetric matrix}
The standard inner product deﬁned between the inertial parameter vector \(\phi_{B_i} \in \Re^{10}\) and a coefficient vector \(s_{B_i} \in \Re^{10}\) can be uniquely restated as the trace
product between pseudo inertia matrix and some \(4\times4\) symmetric matrix as,
\begin{equation*}
    \phi_{B_i}^T\,s_{B_i} = tr(\mathcal{L}_{B_i}\mathcal{S}_{B_i}(s_{B_i}))
\end{equation*}
with 
\begin{equation*}
    S_{B_i}(s) = \begin{bmatrix}
    s_6+s_7 & -0.5s_8 & -0.5s_{10} & 0.5s_2\\
    -0.5s_8 & s_5+s_7 & -0.5s_9 & 0.5s_3 \\
    -0.5s_{10} & -0.5s_9 & s_5+s_6 & 0.5s_4 \\
    0.5s_2 & 0.5s_3 & 0.5s_4 & s_1
    \end{bmatrix}.
\end{equation*}

\subsection{Proof of Theorem 4}
In order to prove Theorem 4, we need to rewrite the first terms of (\ref{equ39}) and (\ref{equ48}) for \(i=1...,7\). It follows from (\ref{equ3}), (\ref{equ7}), (\ref{equ22}), (\ref{equ24}), (\ref{equ48}),
\renewcommand{\theequation}{I.\arabic{equation}}
\setcounter{equation}{0}
\begin{equation}\label{I.1}
\begin{split}
    &(^{B_1}\mathcal{V}_r-\,^{B_1}\mathcal{V})^T(\,^{B_1}\mathcal{F}^*_r -\,^{B_1}\mathcal{F}^*) = (^{B_1}\mathcal{V}_r-\,^{B_1}\mathcal{V})^T\,\\
    &(\,^{B_1}\mathcal{F}_r -\,^{B_1}\mathcal{F}) - (^{B_1}U^T_{T1}\,(^{B_1}\mathcal{V}_r-\,^{B_1}\mathcal{V}))^T\\
    &(\,^{B_1}\mathcal{F}_r -\,^{B_1}\mathcal{F}) = (^{B_1}\mathcal{V}_r-\,^{B_1}\mathcal{V})^T\,
    (\,^{B_1}\mathcal{F}_r -\,^{B_1}\mathcal{F})\\
    &-(^{T_1}\mathcal{V}_r-\,^{T_1}\mathcal{V})^T\,
    (\,^{T_1}\mathcal{F}_r -\,^{T_1}\mathcal{F})=p_{B1}-p_{T1}.
\end{split}
\end{equation}
In the same way, we have for all the frames,
\begin{equation}\label{I.2}
    \begin{split}
        &(^{B_2}\mathcal{V}_r-\,^{B_2}\mathcal{V})^T(\,^{B_2}\mathcal{F}^*_r -\,^{B_2}\mathcal{F}^*) = 
        (^{B_2}\mathcal{V}_r-\,^{B_2}\mathcal{V})^T\,\\
    &(\,^{B_2}\mathcal{F}_r -\,^{B_2}\mathcal{F})
    -(^{T_2}\mathcal{V}_r-\,^{T_2}\mathcal{V})^T\,
    (\,^{T_2}\mathcal{F}_r -\,^{T_2}\mathcal{F})\\
    &=p_{B2}-p_{T2}
    \end{split}
\end{equation}
\begin{equation}\label{I.3}
    \begin{split}
        &(^{B_3}\mathcal{V}_r-\,^{B_3}\mathcal{V})^T(\,^{B_3}\mathcal{F}^*_r -\,^{B_3}\mathcal{F}^*) = 
        (^{B_3}\mathcal{V}_r-\,^{B_3}\mathcal{V})^T\,\\
    &(\,^{B_3}\mathcal{F}_r -\,^{B_3}\mathcal{F})
    -(^{T_3}\mathcal{V}_r-\,^{T_3}\mathcal{V})^T\,
    (\,^{T_3}\mathcal{F}_r -\,^{T_3}\mathcal{F})\\
    &=p_{B3}-p_{T3}
    \end{split}
\end{equation}
\begin{equation}\label{I.4}
    \begin{split}
        &(^{B_4}\mathcal{V}_r-\,^{B_4}\mathcal{V})^T(\,^{B_4}\mathcal{F}^*_r -\,^{B_4}\mathcal{F}^*) = 
        (^{B_4}\mathcal{V}_r-\,^{B_4}\mathcal{V})^T\,\\
    &(\,^{B_4}\mathcal{F}_r -\,^{B_4}\mathcal{F})
    -(^{T_4}\mathcal{V}_r-\,^{T_4}\mathcal{V})^T\,
    (\,^{T_4}\mathcal{F}_r -\,^{T_4}\mathcal{F})\\
    &=p_{B4}-p_{T4}
    \end{split}
\end{equation}
\begin{equation}\label{I.5}
    \begin{split}
        &(^{B_5}\mathcal{V}_r-\,^{B_5}\mathcal{V})^T(\,^{B_5}\mathcal{F}^*_r -\,^{B_5}\mathcal{F}^*) = 
        (^{B_5}\mathcal{V}_r-\,^{B_5}\mathcal{V})^T\,\\
    &(\,^{B_5}\mathcal{F}_r -\,^{B_5}\mathcal{F})
    -(^{T_5}\mathcal{V}_r-\,^{T_5}\mathcal{V})^T\,
    (\,^{T_5}\mathcal{F}_r -\,^{T_5}\mathcal{F})\\
    &=p_{B5}-p_{T5}
    \end{split}
\end{equation}
\begin{equation}\label{I.6}
    \begin{split}
        &(^{B_6}\mathcal{V}_r-\,^{B_6}\mathcal{V})^T(\,^{B_6}\mathcal{F}^*_r -\,^{B_6}\mathcal{F}^*) = 
        (^{B_6}\mathcal{V}_r-\,^{B_6}\mathcal{V})^T\,\\
    &(\,^{B_6}\mathcal{F}_r -\,^{B_6}\mathcal{F})
    -(^{T_6}\mathcal{V}_r-\,^{T_6}\mathcal{V})^T\,
    (\,^{T_6}\mathcal{F}_r -\,^{T_6}\mathcal{F})\\
    &=p_{B6}-p_{T6}
    \end{split}
\end{equation}
\begin{equation}\label{I.7}
    \begin{split}
        &(^{B_7}\mathcal{V}_r-\,^{B_7}\mathcal{V})^T(\,^{B_7}\mathcal{F}^*_r -\,^{B_7}\mathcal{F}^*) = 
        (^{B_7}\mathcal{V}_r-\,^{B_7}\mathcal{V})^T\,\\
    &(\,^{B_7}\mathcal{F}_r -\,^{B_7}\mathcal{F})
    -(^{T_7}\mathcal{V}_r-\,^{T_7}\mathcal{V})^T\,
    (\,^{T_7}\mathcal{F}_r -\,^{T_7}\mathcal{F})\\
    &=p_{B7}-p_{T7}.
    \end{split}
\end{equation}

By subtracting \(\tau^*=\tau_i-\tau_{ai}\) from (\ref{equ26}), one can write,
\begin{equation}\label{I.8}
    \tau^*_{ri}-\tau^*_i = -(\tau_{ari}-\tau_{ai}).
\end{equation}
Then, utilizing (\ref{equ3}), (\ref{equ7}), (\ref{equ17}), (\ref{equ19}), (\ref{I.8}), and (\ref{equ27}), we can write the first term of (\ref{equ48}) as,
\begin{equation}\label{I.9}
    \begin{split}
        &(\Dot{q}_{r1}-\,\Dot{q}_1)(\tau^*_{r1}-\,\tau^*_1) = -(\Dot{q}_{r1}-\,\Dot{q}_1)(\tau_{ar1}-\,\tau_{a1})=\\
        & -((\,^{B_1}\mathcal{V}_r -\,^{B_1}\mathcal{V})
        -^{T_{0}}U_{B_1}^T(\,^{T_0}\mathcal{V}_r -\,^{T_0}\mathcal{V}))^T\\
        &(\,^{B_1}\mathcal{F}_r -\,^{B_1}\mathcal{F}) = 
        -(\,^{B_1}\mathcal{V}_r -\,^{B_1}\mathcal{V})^T(\,^{B_1}\mathcal{F}_r -\,^{B_1}\mathcal{F})\\
        &+
        (\,^{T_0}\mathcal{V}_r -\,^{T_0}\mathcal{V})^T
        (\,^{T_0}\mathcal{F}_r -\,^{T_0}\mathcal{F})= -p_{B1}+p_{T0}.
    \end{split}
\end{equation}

In the same way, it can be done for all the joints,
\begin{equation}\label{I.10}
    \begin{split}
        &(\Dot{q}_{r2}-\,\Dot{q}_2)(\tau^*_{r2}-\,\tau^*_2) = -(\Dot{q}_{r2}-\,\Dot{q}_2)(\tau_{ar2}-\,\tau_{a2})=\\
        &-(\,^{B_2}\mathcal{V}_r -\,^{B_2}\mathcal{V})^T(\,^{B_2}\mathcal{F}_r -\,^{B_2}\mathcal{F})\\&+
        (\,^{T_1}\mathcal{V}_r -\,^{T_1}\mathcal{V})^T
        (\,^{T_1}\mathcal{F}_r -\,^{T_1}\mathcal{F})=-p_{B2}+p_{T1}.
    \end{split}
\end{equation}
\begin{equation}\label{I.11}
    \begin{split}
        &(\Dot{q}_{r3}-\,\Dot{q}_3)(\tau^*_{r3}-\,\tau^*_3) = -(\Dot{q}_{r3}-\,\Dot{q}_3)(\tau_{ar3}-\,\tau_{a3})=\\
        &-(\,^{B_3}\mathcal{V}_r -\,^{B_3}\mathcal{V})^T(\,^{B_3}\mathcal{F}_r -\,^{B_3}\mathcal{F})\\&+
        (\,^{T_2}\mathcal{V}_r -\,^{T_2}\mathcal{V})^T
        (\,^{T_2}\mathcal{F}_r -\,^{T_2}\mathcal{F})=-p_{B3}+p_{T2}.
    \end{split}
\end{equation}
\begin{equation}\label{I.12}
    \begin{split}
        &(\Dot{q}_{r4}-\,\Dot{q}_4)(\tau^*_{r4}-\,\tau^*_4) = -(\Dot{q}_{r4}-\,\Dot{q}_4)(\tau_{ar4}-\,\tau_{a4})=\\
        &-(\,^{B_4}\mathcal{V}_r -\,^{B_4}\mathcal{V})^T(\,^{B_4}\mathcal{F}_r -\,^{B_4}\mathcal{F})\\&+
        (\,^{T_3}\mathcal{V}_r -\,^{T_3}\mathcal{V})^T
        (\,^{T_3}\mathcal{F}_r -\,^{T_3}\mathcal{F})=-p_{B4}+p_{T3}.
    \end{split}
\end{equation}
\begin{equation}\label{I.13}
    \begin{split}
        &(\Dot{q}_{r5}-\,\Dot{q}_5)(\tau^*_{r5}-\,\tau^*_5) = -(\Dot{q}_{r5}-\,\Dot{q}_5)(\tau_{ar5}-\,\tau_{a5})=\\
        &-(\,^{B_5}\mathcal{V}_r -\,^{B_5}\mathcal{V})^T(\,^{B_5}\mathcal{F}_r -\,^{B_5}\mathcal{F})\\&+
        (\,^{T_4}\mathcal{V}_r -\,^{T_4}\mathcal{V})^T
        (\,^{T_4}\mathcal{F}_r -\,^{T_4}\mathcal{F})=-p_{B5}+p_{T4}.
    \end{split}
\end{equation}
\begin{equation}\label{I.14}
    \begin{split}
        &(\Dot{q}_{r6}-\,\Dot{q}_6)(\tau^*_{r6}-\,\tau^*_6) = -(\Dot{q}_{r6}-\,\Dot{q}_6)(\tau_{ar6}-\,\tau_{a6})=\\
        &-(\,^{B_6}\mathcal{V}_r -\,^{B_6}\mathcal{V})^T(\,^{B_6}\mathcal{F}_r -\,^{B_6}\mathcal{F})\\&+
        (\,^{T_5}\mathcal{V}_r -\,^{T_5}\mathcal{V})^T
        (\,^{T_5}\mathcal{F}_r -\,^{T_5}\mathcal{F})=-p_{B6}+p_{T5}.
    \end{split}
\end{equation}
\begin{equation}\label{I.15}
    \begin{split}
        &(\Dot{q}_{r7}-\,\Dot{q}_7)(\tau^*_{r7}-\,\tau^*_7) = -(\Dot{q}_{r7}-\,\Dot{q}_7)(\tau_{ar7}-\,\tau_{a7})=\\
        &-(\,^{B_7}\mathcal{V}_r -\,^{B_7}\mathcal{V})^T(\,^{B_7}\mathcal{F}_r -\,^{B_7}\mathcal{F})\\&+
        (\,^{T_6}\mathcal{V}_r -\,^{T_6}\mathcal{V})^T
        (\,^{T_6}\mathcal{F}_r -\,^{T_6}\mathcal{F})=-p_{B7}+p_{T6}.
    \end{split}
\end{equation}

Now, the general accompanying function can be defined as,
\begin{equation}\label{I.16}
    \nu(t) = \sum_{i=1}^{7} (\nu_i(t)+\nu_{ai}(t)).
\end{equation}
By taking the time derivative of \mbox{(\ref{I.16})} and using \mbox{(\ref{equ39})}, \mbox{(\ref{equ48})} one can obtain,
\begin{equation}\label{I.17}
\begin{split}
        \Dot{\nu}(t) &= \sum_{i=1}^{7} [(^{B_i}\mathcal{V}_r-\,^{B_i}\mathcal{V})^T(\,^{B_i}\mathcal{F}^*_r -\,^{B_i}\mathcal{F}^*)  \\
        &-(^{B_i}\mathcal{V}_r-\,^{B_i}\mathcal{V})^T\,\mathcal{K}_{Di}(\,^{B_i}\mathcal{V}_r-\,^{B_i}\mathcal{V})\qquad  \\
        &+(\Dot{q}_{ri}-\,\Dot{q}_i)(\tau^*_{ri}-\,\tau^*_i)-k_{di}(\Dot{q}_{ri}-\,\Dot{q}_i)^2].
    \end{split}
\end{equation}
Then, by expanding the summation and replacing \mbox{(\ref{I.1})-(\ref{I.7})} and \mbox{(\ref{I.9})-(\ref{I.15})} in \mbox{(\ref{I.17})} for \mbox{\(i = 1...7\)}, one can achieve,
\begin{equation}\label{I.18}
\begin{split}
        \Dot{\nu}(t) &= \sum_{i=1}^{7} [-(^{B_i}\mathcal{V}_r-\,^{B_i}\mathcal{V})^T\,\mathcal{K}_{Di}(\,^{B_i}\mathcal{V}_r-\,^{B_i}\mathcal{V})\qquad  \\
        &-k_{di}(\Dot{q}_{ri}-\,\Dot{q}_i)^2] +p_{B1}-p_{T1}+p_{B2}-p_{T2}\\
        &+p_{B3}-p_{T3}+p_{B4}-p_{T4}+p_{B5}-p_{T5}+p_{B6}\\
        &-p_{T6}+p_{B7}-p_{T7}-p_{B1}+p_{T0}-p_{B2}+p_{T1}\\
        &-p_{B3}+p_{T2}-p_{B4}+p_{T3}-p_{B5}+p_{T4}-p_{B6}\\
        &+p_{T5}-p_{B7}+p_{T6}.
    \end{split}
\end{equation}
We can see that all the VPFs cancel out each other, and we have,
\begin{equation}\label{I.19}
\begin{split}
        \Dot{\nu}(t) &= \sum_{i=1}^{7} [-(^{B_i}\mathcal{V}_r-\,^{B_i}\mathcal{V})^T\,\mathcal{K}_{Di}(\,^{B_i}\mathcal{V}_r-\,^{B_i}\mathcal{V})\qquad  \\
        &-k_{di}(\Dot{q}_{ri}-\,\Dot{q}_i)^2] +p_{T0}-p_{T7}\\
    \end{split}
\end{equation}
The remaining terms in \mbox{(\ref{I.19})}, \mbox{\(p_{T7}\)} and \mbox{\(p_{T0}\)}, correspond to interaction with an environment and moving ground, respectively. Because of the fact that in our case, the exoskeleton has a fixed base and no contact with the environment, these terms vanish. Therefore, we will have,
\begin{equation}\label{I.20}
\begin{split}
    \Dot{\nu}(t) = -\sum_{i=1}^{7} (k_{di}(\Dot{q}_r-\,\Dot{q}_i)^2+(^{B_i}\mathcal{V}_r-\,^{B_i}\mathcal{V})^T\,\mathcal{K}_{Di}\\
    (\,^{B_i}\mathcal{V}_r-\,^{B_i}\mathcal{V}))\leq 0.
\end{split}
\end{equation}
Thereby, the asymptotic stability of the entire system under the designed decentralized controller and considering Theorem 1 and \mbox{(\ref{I.20})} is achieved. Consequently, it can be concluded that for a given bounded desired trajectory \mbox{\(q_d\)}, \mbox{\(\Dot{q}_d\)}, and \mbox{\(\Ddot{q}_d\)}, the boundedness of all the signals is ensured. Moreover, with respect to Lemma 2 and \mbox{(\ref{I.20})}, it can be concluded that \mbox{\(|\mathbf{e}_a|<k_b\)} for \mbox{\(t>0\)}, which ensures joint will stay within a safe region without constraint violation.

\bibliographystyle{IEEEtran}
\bibliography{mybibfile}

\vfill

\end{document}